\documentclass[reviewcopy]{elsart}

\usepackage[utf8]{inputenc}
\usepackage{graphicx}
\usepackage{amsmath}
\usepackage{amssymb}
\usepackage{amsfonts}
\usepackage{subfigure}
\usepackage{ifthen}
\usepackage{wasysym}
\usepackage{color}
\usepackage{url}
\usepackage{lastpage}
\usepackage[T1]{fontenc}
\usepackage{ae,aecompl}
\usepackage{mathrsfs}
\usepackage{multirow}

\journal{Acta Astronautica}

\def\venere   {\leavevmode\raise0.2ex\hbox{\wasyfamily\char25}}
\def\marte    {\leavevmode\lower0.2ex\hbox{\wasyfamily\char26}}

\newcommand{\define}{\triangleq}                                

\newcommand{\V}[1]{\ensuremath{\boldsymbol{#1}}}                
\newcommand{\T}{\textnormal{\tiny{T}}} 

\newcommand{\rev}[1]{\color{black} #1 \color{black}}

\newcommand{\diff}{\textnormal{d}}

\newcommand{\unit}[1]{\,\mathrm{#1}}  
\renewcommand{\sp}{\textnormal{sp}}
\renewcommand{\max}{\textnormal{max}}

\newcommand{\xoe}{\textbf{\textit{\oe}}}
\allowdisplaybreaks[4]

\begin{document}

\begin{frontmatter}

\title{\textsf{Analytical Shaping Method for Low-Thrust Rendezvous Trajectory Using Cubic Spline Functions}}

\begin{center}

\author{Di Wu}, $\;$%
\ead{wud17@mails.tsinghua.edu.cn}\corauth[cor1]{{Corresponding author}.}
\author{Tongxin Zhang}, $\;$%
\ead{ztx18@mails.tsinghua.edu.cn}
\author{Yuan Zhong}, $\;$%
\ead{y-zhong17@mails.tsinghua.edu.cn}
\author{Fanghua Jiang\corauthref{cor1}}, $\;$%
\ead{jiangfh@tsinghua.edu.cn}
\author{Junfeng Li}
\ead{lijunf@tsinghua.edu.cn}

\address{School of Aerospace Engineering, Tsinghua University, 100084 Beijing, People's Republic of China}

\end{center}

\begin{abstract}
\begin{sffamily}
Preliminary mission design requires an efficient and accurate approximation to the low-thrust rendezvous trajectories, which might be generally three-dimensional and involve multiple revolutions. In this paper, a new shaping method using cubic spline functions is developed for the analytical approximation, \rev{which} shows advantages in the optimality and computational efficiency. The rendezvous constraints on the boundary states and transfer time are all satisfied analytically, under the assumption that the boundary conditions and segment numbers of cubic spline functions are designated in advance. Two specific shapes are then formulated according to whether they have free optimization parameters. The shape without free parameters provides an efficient and robust estimation, while the other one allows a subsequent optimization for the satisfaction of additional constraints such as the \rev{constraint on the thrust magnitude.} Applications of the proposed method in combination with the particle swarm optimization algorithm are discussed through two typical interplanetary rendezvous missions, \rev{that is,} an inclined multi-revolution trajectory from the Earth to asteroid Dionysus and a multi-rendezvous trajectory of sample return. Simulation examples show that the proposed method is superior to existing methods in terms of providing good estimation for the global search and generating suitable initial guess for the subsequent trajectory optimization.
\end{sffamily}
\end{abstract}


\end{frontmatter}

{\section*{Nomenclature}}%
\vspace{0.3cm}
\begin{tabbing}
	XXXXXXXXXXXXX \= XX \= \kill
	$\V{a}$                                         \> $=$ \>   acceleration vector, [$\unit{m/s^2}$] \\
	$\V{a}_i, \, \V{b}_i, \, \V{c}_i, \, \V{d}_i$   \> $=$ \>   shaping polynomial coefficients of $i$-th segment \\
	$a_t, \, b_t, \, c_t$                           \> $=$ \>   polynomial coefficients for transfer time constraint, [$\unit{s}$] \\
	$\V{A}, \V{B}$                                  \> $=$ \>   matrices in motion equations \\
	$c$                                             \> $=$ \>   number of discrete constrained points \\ 
	$C^2$                                           \> $=$ \>   set of second-order continuous functions \\ 
	$g_0$                                           \> $=$ \>   standard acceleration of gravity, [$\unit{m/s^2}$] \\
	$I_{\rm sp}$                                    \> $=$ \>   specific impulse, [$\unit{s}$] \\
	$J$                                             \> $=$ \>   performance index \\	
	$m$                                             \> $=$ \>   spacecraft mass, [$\unit{kg}$] \\
	$n$                                             \> $=$ \>   number of finite segments \\
	$\V{r}, \V{v}$                                  \> $=$ \>   spacecraft position and velocity, [$\unit{m}$ and $\unit{m/s}$] \\
	$\mathbb{R}$                                    \> $=$ \>   set of all real numbers \\
	$s$                                             \> $=$ \>   normalized independent variable \\
	$S$                                             \> $=$ \>   domain of independent variable \\
	$t$                                             \> $=$ \>   coordinate time, [$\unit{s}$] \\
	$T_{\max}$                                      \> $=$ \>   maximum thrust magnitude, [$\unit{N}$] \\
	$\V{u}$                                         \> $=$ \>   control vector, [$\unit{m/s^2}$] \\
	$\V{x}$                                         \> $=$ \>   state vector \\
	$\V{z}$                                         \> $=$ \>   undetermined parameters \\
	$\Delta_t$					 \> $=$ \>   discriminant for the transfer time constraint \\
	$\Delta_p$					 \> $=$ \>   discriminant for the semi-latus rectum constraint \\ 
	$\phi$                       \> $=$ \>   boundary state cost function \\
	$\varphi$                    \> $=$ \>   running cost function \\
	$\mu$                        \> $=$ \>   gravitational parameter, [$\unit{m^3/s^2}$] \\
	$\rho$                       \> $=$ \>   penalty function \\
	$\V{\xi}$                    \> $=$ \>   shape function \\
	$\hbar$                      \> $=$ \>   magnitude of angular momentum \\
	$\xoe = \left[p,\, f,\, g,\, h,\, k,\, L\right]^{\T}$              \> $=$ \>   modified equinoctial orbital elements \\
\end{tabbing}

\vspace{-0.2cm}
\noindent \emph{\underline{Subscripts}}

\begin{tabbing}
	XXXXXXXXXXXXX \= XX \= \kill
	$_f$                          \> $=$ \>  final \\
	$_{\max}$                     \> $=$ \>  maximum \\
	$_{\min}$                     \> $=$ \>  minimum \\
	$_{\rm opt}$                  \> $=$ \>  optimal \\
	$_r, \, _{\theta}, \, _h$     \> $=$ \>  components in the radial, in-track, and cross-track directions \\
	$_0$                          \> $=$ \>  initial \\
\end{tabbing}

\vspace{-0.2cm}
\noindent \emph{\underline{Superscripts}}

\begin{tabbing}
	XXXXXXXXXXXXX \= XX \= \kill
	$^{-1}$                                          \> $=$ \>  inverse \\
	$^{\left(1\right)}, \, ^{\left(2\right)}$        \> $=$ \>  the first and second solution \\
	$^{\textnormal{T}}$                              \> $=$ \>  transpose \\
	$\cdot$                                          \> $=$ \>  time derivative \\
	$^\prime$                                        \> $=$ \>  derivative to shaping independent variable \\
	
\end{tabbing}

\section{Introduction}
In the past few decades, the low-thrust propulsion has gained great interest for its \rev{specific impulse higher than that of traditional chemical propulsion \cite{zuiani2012direct,topputo2014survey} and almost no fuel consumption when using the innovative solar balloons \cite{BASSETTO2021451} and the electric solar sails \cite{bassetto2021electric}.} The applications have been sufficiently validated in \rev{some} practical missions, \rev{such as} the Hayabusa 2 \cite{kawaguchi2008hayabusa} and BepiColombo \cite{benkhoff2010bepicolombo}. Numerous methods have been proposed for the global search and local optimization of the continuous low-thrust trajectory \cite{betts1998survey,cheng2020real,conway2012survey}. The global search (or preliminary design) usually find the parameters that are critical to the overall performance over a broad search space, for example the rendezvous or gravity-assist sequences and epochs in the deep-space mission scenarios \cite{chen2018multi,petropoulos2004shape}. By comparison, the local optimization solves the optimal control problem based on the results of the global search \cite{petropoulos2004shape}, generating the solution that satisfies the first-order necessary conditions (i.e., the Karush-Kuhn-Tucker condition \cite{song2021adaptive} for the direct method). Obviously, accurate evaluation of the local solution can help the global search to find the global optimal solution, which require the local optimal solution algorithm to be extremely fast.

For the general case of low-thrust trajectory in a central gravitational field, the motion equations are not analytically integrable \cite{jiang2017improving,li2019interplanetary} and the optimal control problem should be numerically solved by the direct or indirect methods, which are admittedly time-consuming to \quad conduct \cite{betts1998survey,Warm2021Di}. So far, \rev{much effort has been devoted to the approximate or exact analytical solutions for some specific thrust profiles (e.g., the constant tangential thrust \cite{boltz1992orbital,bassetto2021generalized} and radial thrust \cite{prussing1998constant,bassetto2021spiral})} and for the particular transfers between near-circular orbits \cite{edelbaum1961propulsion,wen2021low}. These solutions show the effectiveness in fast estimation, but may be far away from the optimal trajectory because their assumptions are only reasonable for limited cases. To approximate the solution with large eccentricity or inclination, some shape-based methods are proposed, \rev{such as} the exponential sinusoid and logarithmic spiral shapes for the low-thrust interception \cite{petropoulos2004shape,bassetto2021spiral}, the inverse polynomial and finite Fourier series shapes for the near-coplanar transfer and rendezvous \cite{wall2009shape,taheri2016initial}, and the pseudo-equinoctial elements (not including the true longitude) and spherical shapes for the three-dimensional (3-D) problems \cite{novak2011improved}. These methods are used extensively for the mission design in combination with some global search algorithms \cite{caruso2020shape}. However, there are still some common problems to be studied further. \rev{One is the shaping optimality for the 3-D multi-revolution rendezvous \cite{zeng2017shape}, which has been discussed in \cite{VASILE2007on} and evaluated by solving an equivalent optimal control problem. Moreover, the optimality can be assessed through taking the shape solutions as the initial guesses of the direct and indirect methods \cite{jiang2017improving,VASILE2007on}. Compared with the coplanar case, it is more challenging to design the near-optimal solutions of the 3-D trajectories because of their complexity.} Moreover, the satisfaction of the rendezvous time constraint may lead to a nonlinear equation that can be solved by the Newton's method. Since the Newton's method may \rev{fail} to converge when used to solve the nonlinear equation, the robustness of the shaping algorithm cannot be ensured. The spacecraft and engine's physical parameters (e.g., the initial mass, the thrust magnitude, and the specific impulse) are rarely considered by most of the existing shape-based methods, resulting in that the shape-based trajectory may be infeasible considering the practical constraints.

The aim of this paper is to provide an analytical shaping method by using the cubic spline functions to represent the slow variables of a general 3-D multi-revolution rendezvous trajectory. The idea of shaping the magnitude of the angular momentum is introduced, such that the rendezvous time constraint can be solved analytically. A rapid and robust shaping method is then formulated by suitably designating the segment numbers of cubic spline functions. In addition, by adding free interior points, the optimal control problem is transformed into a parameter optimization problem, where the practical constraints on the spacecraft and engine's parameters can be considered to avoid the infeasible solution. Compared with the existing method \cite{zeng2017shape}, the proposed method is superior in terms of the optimality and thrust magnitude, and these \rev{characteristics} are further improved by the parameter optimization process. Therefore, the aforementioned problems are satisfactorily overcome by the proposed method. Numerical examples show the superiority of the proposed method, not only in providing an efficient and robust approximation for global search, but also in generating the suitable initial guess for the local optimization.

The rest of this paper is organized as follows. In Sec.~2, the problem of shaping a low-thrust trajectory with parameterized shape functions is introduced, and a parameter optimization problem is then formulated for the corresponding optimal control problem. In Sec.~3, the formulation of shaping trajectory with cubic spline functions is presented, and two types of shaping methods are proposed and compared according to whether they have free optimization parameters. The applications of the proposed trajectory design method are demonstrated by two specific missions in Sec.~4, and Section~5 concludes this paper.

\section{Problem Formulation for Shaping Low-Thrust Trajectory}

The problem considered here is to approximate the low-thrust trajectory with a near-optimal shape function. The function is intentionally designed to satisfy some constraints, such that a fast estimation for the mission design or an elegant initial guess for the subsequent trajectory optimization can be obtained. It is usually chosen to shape the state vector to avoid the numerical integration of the low-thrust motion equations, and its derivatives are used to analytically evaluated the control law. The general 3-D motion equation for the spacecraft propelled by the low thrust can be formulated as \cite{novak2011improved}
\begin{equation}
\label{dyn}
\dot{\V{x}} = \V{A}\left(\V{x},\,t\right) + \V{B}\left(\V{x},\,t\right) \, \V{u}
\end{equation}
where $\V{x} \in \mathbb{R}^{6}$ is the state vector, $\V{u} \in \mathbb{R}^{3}$ is the thrust acceleration, and the expressions of $\V{A} \in \mathbb{R}^6$ and $\V{B} \in \mathbb{R}^{6\times3}$ depend on the forces acting on the spacecraft and the \rev{considered coordinate system.} The 3-D position $\V{r}$ and velocity $\V{v}$ of the spacecraft are given by the state $\V{x}$ through a transformation from the employed coordinate to the Cartesian coordinate. The thrust acceleration $\V{u}$ is obtained by an inverse transformation \rev{of} Eq.~\eqref{dyn}:
\begin{equation}
\label{dyn_u}
\V{u} = \left[\V{B}\left(\V{x},\,t\right)^{\T}\V{B}\left(\V{x},\,t\right)\right]^{-1}\V{B}\left(\V{x},\,t\right)^{\T}\left(\dot{\V{x}} - \V{A}\left(\V{x},\,t\right)\right)
\end{equation} 
where the state $\V{x}$ and its derivative $\dot{\V{x}}$ are given by the shape function. Based on Eq.~\eqref{dyn_u}, the control law $\V{u}$ can always be solved if the thrust is unconstrained, i.e., $\V{B}\left(\V{x},\,t\right)^{\T}\V{B}\left(\V{x},\,t\right)$ is invertible \cite{novak2011improved}. In addition, an implicit condition for a feasible shape trajectory is that the position and velocity given by the shape must satisfy $\dot{\V{r}} = \V{v}$. On this condition, the shape trajectory and the control law can be derived from each other by Eqs.~\eqref{dyn} and~\eqref{dyn_u}, respectively.

Consider a shape function for the position of \rev{the} spacecraft $\V{r} = \V{r}\left(s, \, \V{z}\right)$, where $s$ is an independent variable nominalized in the interval $\left[0,\,1\right]$, and $\V{z}$ is the undetermined parameters designed to satisfy constraints. The velocity of spacecraft is then derived as
\begin{equation}
\label{dyn_v}
\V{v} = \dot{\V{r}} = \dot{s} \, \V{r}^{\prime}
\end{equation} 
where $\left(*\right)^{\prime}$ denotes \rev{a} derivative with respect to $s$ and $t^{\prime} = 1 \,/\, \dot{s}$ is determined by another shape function $t^{\prime} \left(s, \, \V{z}\right)$. For the low-thrust rendezvous trajectory in a central gravitational field, the motion equation is given by
\begin{equation}
\label{dyn_r}
\ddot{s} \, \V{r}^{\prime} + \dot{s}^2 \, \V{r}^{\prime\prime} = -\frac{\mu}{r^3}\V{r} + \V{u}
\end{equation}
where $\mu$ is the gravitational parameter \rev{of the primary body,} $\dot{s} = 1 \,/\, t^{\prime}$, and $\ddot{s} = -t^{\prime\prime} \,/\, t^{\prime \, 3}$. The \rev{spacecraft engine} has a constant specific impulse $I_{\rm sp}$ and a maximum thrust magnitude $T_{\max}$, \rev{therefore} the mass of spacecraft is governed by
\begin{equation}
\label{dyn_m}
	\dot{m} = -\frac{m \, \left\|\V{u}\right\|}{I_{\rm sp}\,g_0}
\end{equation}
where $g_0 = 9.80665 \, \unit{m/s^2}$ is the \rev{standard gravity,} while the thrust magnitude for the shape trajectory is constrained:
\begin{equation}
\label{tmax_con}
	m \, \left\|\V{u}\right\| \leq T_{\max}
\end{equation}
In addition, the rendezvous boundary states (composed of the initial state $\left[\V{r}_0^{\T}, \V{v}_0^{\T}\right]^{\T}$ and \rev{the} final state $\left[\V{r}_f^{\T}, \V{v}_f^{\T}\right]^{\T}$) and \rev{the} transfer time are both fixed. The corresponding constraints are summarized as follows:
\begin{align}
&\V{r}\left(0, \, \V{z}\right) = \V{r}_0, \quad \V{r}^{\prime}\left(0, \, \V{z}\right) /\, t^{\prime}\left(0, \, \V{z}\right)   = \V{v}_0, \label{con_r1} \\[0.3cm] 
&\V{r}\left(1, \, \V{z}\right) = \V{r}_f, \quad \V{r}^{\prime}\left(1, \, \V{z}\right) /\, t^{\prime}\left(1, \, \V{z}\right) = \V{v}_f, \label{con_r2} \\[0.3cm]
&\qquad\; t_f - t_0 = \int_0^1 t^{\prime}\left(s, \, \V{z}\right) \, \diff s \label{con_t}
\end{align}
where $t_0$ and $t_f$ denote the initial and final times, respectively. The shape functions incorporating undetermined parameters (i.e., $\V{r}\left(s, \, \V{z}\right)$ and $t^{\prime}\left(s, \, \V{z}\right)$) are optimized, such that the corresponding trajectory satisfies \rev{the path constraint~\eqref{tmax_con} and boundary constraints~\eqref{con_r1}--\eqref{con_t}.} The trajectory can be determined by these functions, \rev{while the required thrust acceleration and the spacecraft mass} are then obtained \rev{from} Eqs.~\eqref{dyn_r} and~\eqref{dyn_m}, respectively.

Generally, the optimal control problem for the low-thrust trajectory optimization can be transformed into a parameter optimization problem by shaping the trajectory. The performance index for this problem is formulated as
\begin{equation}
\label{J}
J = \phi \left(\V{z}\right) + \int_{0}^{1} \varphi \left(s, \, \V{z}\right) \, \diff s
\end{equation}
where $\phi\left(\V{z}\right)$ is the boundary state cost function and $\varphi\left(s,\,\V{z}\right)$ is the running cost function \cite{ricciardi2019direct}. The dynamic constraints are satisfied automatically via Eq.~\eqref{dyn_u}, and the path and boundary constraints are collected as Eqs.~\eqref{tmax_con}--\eqref{con_r2}. Note that the minimum number of the undetermined parameters is equal to the number of \rev{equality constraints,} that is, 13 in total. In literature, numerous types of functions for $\V{r}\left(s, \, \V{z}\right)$ and $t^{\prime}\left(s, \, \V{z}\right)$ are proposed with different number of parameters, e.g., the inverse polynomial in the \rev{planar} polar coordinate \cite{wall2009shape}, the shape functions in terms of classic or equinoctial elements \cite{zeng2017shape,de2006preliminary}, and the Fourier series in the cylindrical coordinate \cite{taheri2016initial}. Instead of shaping the position $\V{r}\left(s, \, \V{z}\right)$, Ref.~\cite{gondelach2015hodographic} shapes the velocity with some integrable base functions. The independent variable $s$ is usually chosen from the time or the coordinate parameters, e.g., the polar angle or the true anomaly. Meanwhile, the functions with respect to the time are more complex and usually designed with more optimization para- meters \cite{taheri2016initial}, whereas the functions with respect to the coordinate parameter result in that the transfer time constraint is nonlinear and its analytical satisfaction is difficult \cite{novak2011improved}.

In this paper, the shape functions in terms of the modified equinoctial orbital elements \cite{Rapid2021Di} (MEOEs) $\xoe \left(s, \, \V{z}\right) = \left[p, \, f, \, g, \, h, \, k, \, L\right]^{\T}$ are formulated. The independent variable is the true longitude given by
\begin{equation}
\label{Ls}
 L = L_0 + s\, \Delta L
\end{equation}
where $\Delta L = L_f - L_0$, and $L_0$ and $L_f$ are the fixed initial and final true longitudes, respectively. The position of the spacecraft is obtained via the transformation \cite{betts2000very}: 
\begin{equation}
\V{r} = \frac{r}{\beta^2}\left[\begin{aligned}
&\left(1+\alpha^2\right) \cos L + 2 \, h \, k \sin L \\[0.2cm]
&\left(1-\alpha^2\right) \sin L + 2 \, h \, k \cos L \\[0.2cm]
&2\left(h \sin L - k \cos L \right)
\end{aligned}\right] \label{sr}
\end{equation}
where $r = \left\| \V{r}\right\| = p\,/\left(1+f\cos L + g\sin L\right)$, $\alpha^2 = h^2 - k^2$, and $\beta^2 = 1+ h^2 +k^2$. Then, the velocity and acceleration are derived as
\begin{align}
&\V{v} = \dot{\V{r}} = \frac{\diff \V{r}}{\diff \xoe} \dot{s} \, \xoe^{\prime} \label{sv}, \\[0.3cm]
&\V{a} = \dot{\V{v}} = \frac{\diff \V{r}}{\diff \xoe} \ddot{s} \, \xoe^{\prime} + \dot{s}^2 \left(\frac{\diff \V{r}}{\diff \xoe} \xoe^{\prime\prime} + \xoe^{\prime \T}\frac{\diff^2 \V{r}}{\diff \xoe^2} \xoe^{\prime}\right) \label{sa} 
\end{align}
where the derivatives $\dot{s}$ and $\ddot{s}$ are designated through another shape function for the magnitude of the angular momentum $\hbar \left(s, \, \V{z}\right) = r^2 \, \dot{L}$, viz.
\begin{align}
&\dot{s} = \frac{\dot{L}}{\Delta L} = \frac{\hbar}{r^2 \, \Delta L} \label{sd}, \\[0.3cm]
&\ddot{s} = \frac{\ddot{L}}{\Delta L} = \frac{\dot{s}}{r^2 \, \Delta L} \left(\hbar^{\prime} - \frac{2\, \hbar\, \diff r}{r\,\diff \xoe} \xoe^{\prime}\right) \label{sdd}
\end{align}
The detailed expressions for the derivatives $\diff \V{r} / \diff {\xoe}$, $\diff \V{r}^2 / \diff^2 {\xoe}$, and $\diff {r} / \diff {\xoe}$ will be given in the Appendix. Note that the velocity obtained by Eq.~\eqref{sv} may not be equal to the velocity transformed from the MEOEs; that is, these elements shaped here and known as the pseudo-equinoctial elements \cite{novak2011improved} are not osculating. Finally, a trajectory is designed by the shape function $\V{\xi}\left(s, \, \V{z}\right) = \left[p, \, f, \, g, \, h, \, k, \, \hbar\right]^{\T}$, whose expressions will be formulated in the next section using cubic spline functions, such that all the boundary constraints could be satisfied analytically.

\section{Analytical Shaping with Cubic Spline Functions}

In this section, first, the idea of representing the trajectory $\V{\xi} \left(s, \V{z}\right)$ by cubic spline functions is introduced. These functions are second-order continuous, rendering that the velocity and acceleration are both continuous. Then, the constraints on boundary states and transfer time are satisfied by designing the values of boundary and interior points of the cubic spline, respectively. The residual parameters of the cubic spline with multiple segments can be optimized for the satisfaction of some extra constraints (e.g., the thrust magnitude constraint) or better performance index of the trajectory. According to whether there are free optimization parameters, two types of analytical shaping methods are finally proposed, and \rev{their computational performance} are validated through comparisons to previous studies.

\subsection{Analytical Shaping Formulation}

Let us decompose the domain of the independent variable $s$, denoted by $S$ and with the range $\left[0,\, 1\right]$, into $n$ segments such that
\begin{equation}
\label{domain}
S = \mathop\cup\limits_{i=1}^n S_i\left(s_{i-1}, \, s_i\right)
\end{equation}
where $0 = s_0 < s_1 < \cdots < s_n = 1$. At each segment $S_i$, the function $\V{\xi}\left(s, \, \V{z}\right)$ is designed as cubic polynomial:
\begin{equation}
\label{cubicpoly}
\V{\xi}\left(s, \, \V{z}\right) = \V{a}_i \, s^3 + \V{b}_i \, s^2 + \V{c}_i \, s + \V{d}_i, \quad s \in S_i
\end{equation}
where the vectors $\V{a}_i$, $\V{b}_i$, $\V{c}_i$, and $\V{d}_i$ are the coefficients to be settled. Since the position and velocity of the spacecraft must be continuous, the shaping elements $\xoe$, its derivative $\xoe^{\prime}$, and the magnitude of the angular momentum $\hbar$ are all continuous (see Eqs.~\eqref{sr} and~\eqref{sv}). Besides, the acceleration is usually discontinuous in practice, while it is herein assumed to be continuous to approximate the low-thrust trajectory \cite{wall2009shape}. Thus, the function $\hbar\left(s, \, \V{z}\right)$ should be first-order continuous, and the others are second-order continuous. In order to use a uniform formulation, the function $\V{\xi}\left(s, \, \V{z}\right)$ is designed with second-order continuity ($\V{\xi}\left(s, \, \V{z}\right) \in C^2\left[0, \, 1\right]$); that is, it is a cubic spline function. In this case, the undetermined parameter $\V{z}$ consists of the values of $\V{\xi} \left(s_i\right), \, i=0, \, 1, \, \cdots, \, n$ and two additional boundary conditions of the cubic spline, which are usually set according to the first or second order derivatives to $\V{\xi}$ \cite{behforooz1979end}. The other conditions used to settle the coefficients are derived from the aforesaid continuations between adjacent segments, such that the number of conditions is the same as the number of unknown coefficients. 

To determine the cubic spline functions, the following boundary conditions are considered:
\begin{equation}
\label{boundary0}
\V{\xi}^{\prime} \left(0\right) = 0, \quad \V{\xi}^{\prime} \left(1\right) = 0
\end{equation}
By substituting Eqs.~\eqref{sd} and~\eqref{boundary0} into Eq.~\eqref{sv}, the velocity takes the same form as that of the osculating orbital elements, i.e.,  $\V{v} = \dot{L} \, \diff \V{r} / \diff L$. Thus, the shape orbital elements at boundary points are osculating, and the boundary conditions of rendezvous are satisfied by setting the values of the boundary shape functions with the boundary osculating orbital elements and angular momentum magnitudes, viz.
\begin{align}
&\V{\xi} \left(0\right) = \left[p_0, \, f_0, \, g_0, \, h_0, \, k_0, \, \hbar_0\right]^{\T}, \label{boundary1} \\[0.3cm]
&\V{\xi} \left(1\right) = \left[p_f, \, f_f, \, g_f, \, h_f, \, k_f, \, \hbar_f\right]^{\T} \label{boundary2}
\end{align}
The other variables are $\V{\xi}\left(s_i\right), \, i=1,\,2,\,\cdots, \,n-1$ at the interior points. Meanwhile, the value of semi-latus rectum at the first interior point $p_1$ is designated to satisfy the transfer time constraint, while the other variables are free optimization parameters that need to be optimized. For a given final time $t_f$, the transfer time constraint is
\begin{equation}
\label{boundary3}
\int_0^1\frac{p^2 \, \Delta L}{\hbar \left(1+f\cos L + g\sin L\right)^2} \, \diff s + t_0 - t_f = 0
\end{equation}
Note that the shape function $p\left(s\right)$ is proportional to $p_1$ according to the cubic spline formulation, viz.
\begin{equation}
\label{gamma}
	p\left(s\right) = \gamma_1\left(s\right) p_1 + \gamma_2 \left(s\right)
\end{equation}
Then, taking the parameter $p_1$ as the unknown variable, we can integrated Eq.~\eqref{boundary3} as
\begin{equation}
\label{boundary4}
a_t \, p_1^2 + b_t \, p_1 + c_t = 0
\end{equation}
where the coefficients are
\begin{align}
&a_t = \int_0^1 \frac{\gamma_1^2 \, \Delta L}{\hbar \left(1+f\cos L + g \sin L \right)^2} \, \diff s > 0, \label{integration1} \\[0.3cm]
&b_t = \int_0^1 \frac{2\,\gamma_1 \, \gamma_2 \, \Delta L}{\hbar \left(1+f\cos L + g \sin L \right)^2} \, \diff s, \label{integration2} \\[0.3cm]
&c_t = \int_0^1 \frac{\gamma_2^2 \, \Delta L}{\hbar \left(1+f\cos L + g \sin L \right)^2} \, \diff s + t_0 - t_f \label{integration3} 
\end{align}
The free optimization parameters are solved by a direct optimization algorithm, and these coefficients can be integrated rapidly via quadrature using the trapezoidal rule \cite{wall2009shape}. There are two values for $p_1$:
\begin{equation}
\label{solution}
p_1^{\left(1\right)} = \frac{-b_t + \sqrt{b_t^2 - 4 \, a_t\, c_t}}{2 \, a_t}, \quad p_1^{\left(2\right)} = \frac{-b_t - \sqrt{b_t^2 - 4 \, a_t\, c_t}}{2 \, a_t}
\end{equation}
Because $a_t > 0$, the discriminant $\Delta_t = b_t^2 - 4 \, a_t\, c_t$ is an increasing function with respect to the transfer time $t_f - t_0$. The first solution $p_1^{\left(1\right)}$ is also an increasing function of transfer time, while the second one $p_1^{\left(2\right)}$ is a decreasing function. After setting the optimization parameters, there may be no feasible solution (i.e., $\Delta_t < 0$) when the transfer time is too short. In addition, the semi-latus rectum $p_1$ is checked posteriorly to guarantee its value larger than a given minimum value $p_{1 \, {\rm min}}$. Thus, the \rev{following} condition for shaping a feasible trajectory is obtained:
\begin{equation}
\label{conditions}
\Delta_t \geq 0 \, \wedge \, \Delta_p = \max \left\{p_1^{\left(1\right)}, p_1^{\left(2\right)}\right\} - p_{1 \, {\rm min}} \geq 0
\end{equation}
If this condition is violated, $p_1$ is set to $p_0$ in \rev{the} simulation, and the resulting transfer time cannot satisfy its constraint. Later on, this infeasible solution will be avoided by adding the penalty functions to the performance index. If the two solutions are both feasible, the solution with better performance index is preferred. Note that the first one is more consistent with our trajectory design experience because $p_1^{\left(1\right)}$ increases over the transfer time. 

By designating the aforesaid 13 variables $\V{\xi}\left(0\right)$, $\V{\xi}\left(1\right)$, and $p_1$ to satisfy the analytical boundary equality constraints, the free parameters become $\left\{\V{\xi}\left(s_i\right), \, i=1,\,2,\,\cdots, \,n-1\right\} \setminus \left\{p_1\right\}$. The optimization problem is now to minimize the performance index Eq.~\eqref{J} subject to the path constraint Eq.~\eqref{tmax_con}. In this study, the propellant consumption integrated by Eq.~\eqref{dyn_m} is regarded as the performance index:
\begin{equation}
\label{J1}
	J = m_0 \left[1 - \exp\left({-\frac{1}{I_{\rm sp}\,g_0} \int_0^1 {\left\|\V{u}\right\|} \, \diff s}\right)\right] - \rho_1 \min \left\{\Delta_t, \,  0\right\} - \rho_2 \min \left\{\Delta_p, \, 0 \right\}
\end{equation}
where $\exp$ is the exponential function, and $\rho_1$ and $\rho_2$ are penalty functions for those solutions violating Eq.~\eqref{conditions}. In numerical simulation, setting $\rho_1 = \rho_2 = 1 \times 10^4$ is adequate to avoid the infeasible trajectory. Noted that another equivalent choice is to minimize the total velocity increment $\Delta V = \int_0^1 \left\|\V{u}\right\| \, \diff s$, which is independent of the spacecraft and engine's parameters. Thus, the influence of these parameters on the optimal solution is not reflected in the performance index but included in the path constraint.

\subsection{Shaping Methods for 3-D Orbital Rendezvous}

The shape functions for different shaping elements can employ different number of segments; that is, $\V{n} = \left[n_p, \, n_f, \, n_g, \, n_h, \, n_k, \, n_{\hbar}\right]^{\T}$. The number of segments for the semi-latus rectum $n_p$ must be larger than 2 for the rendezvous trajectory, while the minimum values for the others are 1. Thus, the total number of free parameters is $\left\|\V{n}\right\|_1 - 7$. In this subsection, two types of shaping methods with different segment numbers are formulated according to whether they have free optimization parameters. The first one is to shape the rendezvous trajectory rapidly by setting $n_p = 2$ and $n_f = n_g = n_h = n_k = n_{\hbar} = 1$, where the thrust magnitude constraint is not considered. The second one is to shape the trajectory with thrust magnitude constraint by setting $n_p = n_f = n_g = n_h = n_k = n_{\hbar} = n \geq 2$. For the two shaping methods investigated here, the segment intervals of each group of shaping elements are assumed to be equal:
\begin{equation}
\label{interval}
	s_i - s_{i-1} = 1/n, \quad i=1,\,2,\, \cdots, \, n
\end{equation}

\subsubsection{Rapid shaping method neglecting thrust magnitude constraint}

Let $n_p = 2$ and $n_f = n_g = n_h = n_k = n_{\hbar} = 1$, and the rendezvous trajectory is shaped by the corresponding functions. A two-segment piecewise cubic polynomial is formulated for the semi-latus rectum:
\begin{equation}
\label{rapid_p}
p = \left\{\begin{aligned}
&p_0 + 3\left(p_f - p_0\right) \, s^2 - 2\left(p_f - p_0\right)\, s^3 + 4\left(3\,s^2 - 4\,s^3\right) \Delta p, \qquad\qquad\;\;\;\; \textnormal{if} \; s \leq 0.5 \\[0.2cm]
&p_0 + 3\left(p_f - p_0\right) \, s^2 - 2\left(p_f - p_0\right)\, s^3 - 4\left(1 - 6\,s + 9\,s^2 - 4\,s^3\right) \Delta p, \quad \textnormal{if} \; s > 0.5
\end{aligned} \right.
\end{equation}
where $\Delta p \define p_1 - \left(p_f - p_0\right)/2$. The other elements are also shaped by cubic polynomials:
\begin{align}
& f = f_0 + 3\left(f_f - f_0\right) \, s^2 - 2\left(f_f - f_0\right)\, s^3\label{rapid_f} \\[0.3cm]
& g = g_0 + 3\left(g_f - g_0\right) \, s^2 - 2\left(g_f - g_0\right)\, s^3\label{rapid_g} \\[0.3cm]
& h = h_0 + 3\left(h_f - h_0\right) \, s^2 - 2\left(h_f - h_0\right)\, s^3\label{rapid_h} \\[0.3cm]
& k = k_0 + 3\left(k_f - k_0\right) \, s^2 - 2\left(k_f - k_0\right)\, s^3\label{rapid_k} \\[0.3cm]
& \hbar = \hbar_0 + 3\left(\hbar_f - \hbar_0\right) \, s^2 - 2\left(\hbar_f - \hbar_0\right)\, s^3\label{rapid_hh}
\end{align}
All the boundary constraints discussed in the previous subsection are satisfied by substituting Eqs.~\eqref{rapid_p}--\eqref{rapid_hh} into Eqs.~\eqref{boundary0}--\eqref{boundary3}. To simplify the formulation for the semi-latus rectum, $\Delta p$ is designed to meet the transfer time constraint. A quadratic equation can be obtained and solved similarly \rev{to} Eq.~\eqref{boundary4}. When the value of $\Delta p$ is zero, the piecewise formulation becomes a cubic polynomial function. The integral of Eq.~\eqref{boundary3} can be obtained analytically when the rendezvous trajectory between circular orbits is considered. However, for the general cases of rendezvous trajectories with multiple revolutions and large out-of-plane motions, the trapezoidal rule is employed to integrate the transfer time. 

\subsubsection{Shaping method with thrust magnitude constraint}

By setting $n_p = n_f = n_g = n_h = n_k = n_{\hbar} = n \geq 2$ and optimizing the free parameters $\left\{\V{\xi}\left(s_i\right), \, i=1,\,2,\,\cdots, \,n-1\right\} \setminus \left\{p_1\right\}$,  the shape functions can be formulated through a standard cubic spline interpolation algorithm. In this case, it is difficult to give the explicit expressions of $\gamma_1\left(s\right)$ and $\gamma_2\left(s\right)$. To obtain the coefficients of the quartic equation for the transfer time, three values of $\Delta p \define p_1 - \left(p_0+p_2\right)/2$ are tested, i.e., $\Delta p_1 = \left(p_2-p_0\right)/2, \, \Delta p_2 = -\Delta p_1,$ and $\Delta p_3 = 0$. Their transfer times are denoted by $t_1, \, t_2,$ and $t_3$, respectively. Then, the coefficients are obtained \rev{as}
\begin{equation}
 \label{time_constraint}
 c_t = t_3 + t_0 - t_f, \quad b_t = \frac{t_1 - t_2}{2 \, \Delta p_1}, \quad a_t = \frac{t_1 + t_2 - 2 \, t_3}{2 \, \Delta p_1^2}
\end{equation}
Instead of Eqs.~\eqref{integration1}--\eqref{integration3}, the above equations are utilized to solve the exact $\Delta p$ that satisfies the transfer time constraint.

The thrust magnitude constraint is considered at some discrete points, viz,
\begin{equation}
 \label{thrust_constraint}
 m\left(s_{ij}\right) \left\|\V{u}\left(s_{ij}\right)\right\| \leq T_{\max}, \quad j = 0,\,1,\,\cdots,\,c
\end{equation}
where $s_{i-1} = s_{i0} < s_{i1} < \cdots < s_{ic} = s_{i}, \, i = 1,\, 2,\, \cdots, \, n$, and $c$ is the number of discrete points of segment $S_i$. The number $c \geq 1$ is set to the same value for every segment because their intervals are equal. Assuming that these discrete points are also equally spaced, the value of independent variable at each point is
\begin{equation}
\label{pointvalue}
	s_{ij} = \frac{1}{n}\left(i-1+\frac{j}{c}\right), \quad i=1,\,2,\, \cdots, \, n; \; j=0,\,1,\,\cdots, \, c
\end{equation}
Since the boundary points of adjacent segments are counted twice, the total number of constrained points is $nc + 1$. After setting the values of $n$ and $c$, a parameter optimization algorithm is utilized to find the propellant-optimal solution with thrust magnitude constraint.


\subsection{Computational Comparisons to Previous Studies}

To verify the proposed shaping methods, this subsection presents a rendezvous example, in which the initial and target orbital elements in the heliocentric ecliptic J2000 frame are set to compare the results with those of Ref.~\cite{zeng2017shape}. As shown in Table~\ref{table1}, a general 3-D rendezvous between elliptic orbits is performed. The spacecraft departs from the initial orbit at \rev{$t_0 = 0$} and arrives the target orbit at $t_f$. The astronomical unit is set to $1 \, \unit{AU} = 149597870.7\,\unit{km}$, and the gravitational parameter of the Sun is $\mu = 132712440018 \, \unit{km^3/s^2}$.

%
\begin{table}[!htb]
	\caption{Initial and target classic orbital elements \cite{zeng2017shape}}
	\label{table1}
	\vspace{0.25cm}
	\centering
	\begin{tabular}{lcccccc}
		\hline \hline
		Orbits & $a,\, \unit{AU}$ & $e$ & $i,\,\unit{deg} $ & $\Omega, \, \unit{deg}$ & $\omega, \, \unit{deg}$ & $f, \, \unit{deg}$\\ \hline
		Initial & 1.0 & 0.4 & 10.0 & 15.0 & 25.0 & 10.0 \\
		Target & 3.0 & 0.6 & 40.0 & 25.0 & 25.0 & 40.0\\
		\hline \hline
	\end{tabular}
\end{table}
%

The results of the shape trajectories neglecting the thrust magnitude constraint are firstly presented. As stated before, the trajectory is independent of the parameters of spacecraft and engine, and these parameters will be chosen later on for the shaping method with thrust magnitude constraint. The rendezvous trajectories are shaped with a series of transfer times $t_f = 8, \, 16, \, 24, \, 32, \, 40, $ and $ 48 \, \unit{years}$, in which multiple revolutions are usually employed. The optimal revolution numbers that yield the minimum velocity increments are obtained, and the corresponding velocity increments $\Delta V$ and maximum thrust accelerations $u_{\max}$ during the rendezvous are compared in Table~\ref{table2}. It shows that the rendezvous trajectories generated by the proposed method take smaller velocity increments and thrust accelerations. The computational results of the proposed method are more superior than the previous results for all test cases. \rev{The computational efficiency will be demonstrated later on in Sec. 4.1.} As the transfer time increases, the trajectory needs more revolution number, larger velocity increment, and smaller thrust acceleration.

%
\begin{table}[!htb]
	\caption{Comparisons of rendezvous results with different transfer times}
	\label{table2}
	\vspace{0.25cm}
	\centering
	\begin{tabular}{ccccccc}
		\hline \hline
		\multirow{2}{*}{$t_f, \, \unit{years}$} & \multicolumn{3}{c}{Proposed Shaping Method} & \multicolumn{3}{c}{Shaping Method in Ref.~\cite{zeng2017shape}} \\
		& $N_{\rm opt}$ & $\Delta V, \, \unit{km/s}$ & $u_{\max}, \, \unit{mm/s^2} $ & $N_{\rm opt}$ & $\Delta V, \, \unit{km/s}$ & $u_{\max}, \, \unit{mm/s^2} $ \\ \hline
		8  & 3  & 23.01 & 1.22 & 3  & 27.77 & 2.41 \\
		16 & 6  & 22.66 & 0.64 & 5  & 31.25 & 1.74\\
		24 & 9  & 23.29 & 0.44 & 7  & 34.66 & 1.35 \\
		32 & 12 & 24.69 & 0.35 & 9  & 38.91 & 1.17 \\
		40 & 15 & 26.67 & 0.29 & 11 & 43.64 & 1.09 \\
		48 & 18 & 29.07 & 0.25 & 12 & 48.23 & 0.97 \\
		\hline \hline
	\end{tabular}
\end{table}
%

The rendezvous case with $t_f = 16 \, \unit{years}$ is then tested to demonstrate the effectiveness of the proposed shaping method with the thrust magnitude constraint. The spacecraft is assumed to have an initial mass $m_0 = 4000 \, \unit{kg}$, a constant specific impulse $I_{\sp} = 3000 \, \unit{s}$, and a maximum thrust magnitude $T_{\max} = 0.6 \, \unit{N}$. The initial thrust acceleration is $T_{\max}/m_0 = 0.15 \, \unit{mm/s^2}$, which is admittedly smaller than the obtained result $u_{\max} = 0.8 \, \unit{mm/s^2}$ (shown in Table~\ref{table2}). The Powell's COBYLA algorithm \cite{powell1994direct} in the NLopt library is used for the parameter optimization, where the relative tolerance is set to $10^{-4}$. The first guesses for the free parameters are generated by the rapid shaping results. The computational results with different numbers of segments and discrete points are listed in Table~\ref{table3}. It is shown that better performance is obtained as the segment number increases. The thrust magnitude constraint is more strictly satisfied when more discrete points are set, and the corresponding performance index becomes worse in general. It is a trade-off between the performance index and the thrust magnitude constraint. Nevertheless, all results obtained here are better than the results shown in Table~\ref{table2}, but at the cost of more computational time for the parameter optimization. \rev{Note that the COBYLA algorithm uses a linear approximation to the constraints and performance index, whose computational efficiency is admittedly poor for the nonlinear optimization problem. As shown in Table~\ref{table3}, the optimization takes a few minutes to obtain the rendezvous trajectories with the thrust magnitude constraint. Thus, to achieve fast estimation, the analytical shaping method neglecting the thrust magnitude constraint or optimization with larger relative tolerance is suggested.} 

%
\begin{table}[!htb]
	\caption{Comparisons of rendezvous results with thrust magnitude constraint}
	\label{table3}
	\vspace{0.25cm}
	\centering
	\begin{tabular}{ccccccc}
		\hline \hline
		Solution & $n$ & $c$ & $T_{\max}, \, \unit{N} $ & $\Delta V, \, \unit{km/s}$ & $m_{f}, \, \unit{kg} $ & \rev{CPU time, min} \\ \hline
		1 & 10 & 5  & 0.73135 & 20.72 & 1978.06 & \rev{3.84} \\
		2 & 10 & 10 & 0.60836 & 20.90 & 1966.08 & \rev{4.10} \\
		3 & 15 & 5  & 0.61826 & 17.24 & 2225.97 & \rev{11.40} \\
		4 & 15 & 10 & 0.60734 & 17.35 & 2217.75 & \rev{5.25} \\
		5 & 20 & 5  & 0.60687 & 16.28 & 2299.92 & \rev{10.99} \\
		6 & 20 & 10 & 0.60193 & 16.27 & 2300.62 & \rev{10.83} \\
		\hline \hline
	\end{tabular}
\end{table}
%

The proposed shaping method shows its effectiveness in providing fast and accurate estimation for the preliminary mission design. Compared with the results reported in Ref.~\cite{zeng2017shape}, the results generated by the proposed method have better performance, and the efficiency and robustness of shaping are improved because no Newtonian iterative process is required. In addition, the proposed shaping method may not achieve the optimal solution even it uses many segments, due to the subjective assumptions of second-order continuity and zero boundary conditions (see Eq.~\eqref{boundary0}). Thus, to find the optimal trajectory, we should use the proposed method to find good initial guesses and then turn to the traditional direct or indirect methods \cite{topputo2014survey,yang2018fast,zeng2017searching,wang2020asteroid}.

\section{Applications to Trajectory Design}

The shaping method has been widely used to approximate the low-thrust transfer, rendezvous, and gravity-assist trajectories in combination with the global search algorithms \cite{wall2009shape,caruso2020shape}. However, the performance of trajectory design is highly dependent on the estimation accuracy of the shape functions. For the 3-D multi-revolution rendezvous trajectories, the applications of proposed shaping method are presented in this section through two mission scenarios, a rendezvous mission from the Earth to asteroid Dionysus \cite{taheri2018generic} and a Near-Earth Asteroids sample return mission \cite{chen2018multi}. The epochs of these two missions are designed by using the proposed shaping method for estimation and the particle swarm optimization (PSO) method \cite{wu2018problem} for global search. \rev{To achieve the fast estimation, the proposed shaping method neglecting the thrust magnitude constraint is employed here.} Two other estimation methods are compared: 1) the impulsive solution to the Lambert problem \cite{chen2018multi}; 2) the shaping method reported in Ref.~\cite{zeng2017shape}. Based on the PSO search results, the final propellant-optimal trajectories for the two missions are solved by the indirect method. The performance of the proposed method in initial guessing is presented with comparison to the results in Ref.~\cite{jiang2017improving}.

%
\begin{table}[!htb]
	\caption{Classic orbital elements of the Earth and targets (56000 MJD)}
	\label{table4}
	\vspace{0.25cm}
	\centering
	\begin{tabular}{ccccccc}
		\hline \hline
		Name & $a,\, \unit{AU}$ & $e$ & $i,\,\unit{deg} $ & $\Omega, \, \unit{deg}$ & $\omega, \, \unit{deg}$ & $f, \, \unit{deg}$\\ \hline
		Earth & 0.999584 & 0.016375 & 0.002666 & 134.239190 & 329.982886 & 69.425162 \\
		Dionysus & 2.199238 & 0.541127 & 13.526692 & 82.074057 & 204.296334 & 180.509774 \\
		1999 $\textnormal{AO}_{10}$ & 0.911569 & 0.110968 & 2.624497 & 313.313332 & 7.678286 & 186.819009 \\
		2000 $\textnormal{LG}_{6}$ & 0.917259 & 0.110893 & 2.830149 & 72.571729 & 8.144765 & 312.238767 \\
		\hline \hline
	\end{tabular}
\end{table}
%

\subsection{Rendezvous with Dionysus}

Consider a multi-revolution rendezvous mission, in which the spacecraft departs from the Earth to rendezvous with the asteroid Dionysus. The orbital elements of the Earth and Dionysus are listed in Table~\ref{table4}. The spacecraft is assumed to have a constant specific impulse $I_{\sp} = 3000 \, \unit{s}$, a maximum thrust magnitude $T_{\max} = 0.32 \, \unit{N}$, and an initial mass $m_0 = 4000\, \unit{kg}$. The launch epoch $t_0$ is chosen in $t_0 \in \left[56000,\, 56500\right] \, \unit{MJD}$, and the arrival epoch $t_f$ is in $t_f \in \left[59000, \, 60000\right] \, \unit{MJD}$. Then, the PSO algorithm is used to optimize the launch and arrival epochs. Because multiple revolutions are needed, the optimal revolution numbers that take the minimum propellant consumption are chosen identically for all the estimation methods.

The results of epochs are collected in Table~\ref{table5}. The maximal iteration number for the PSO algorithm is set to $n_{\max} = 100$, and the swarm size is set to $S = 20$. The simulation programs are written in C++, compiled with Microsoft Visual Studio 2012 using the release mode and single thread, and run on a desktop computer with an Intel Core i7-7700 CPU of 3.6 GHz and 8.00 GB of RAM. Because the Lambert problem uses two impulsive maneuvers to perform the rendezvous, its estimated propellant consumption is smallest. The solving of the Lambert problem is much faster than the other methods. After the first impulse is applied, the spacecraft enters a transfer orbit, and the value of its semi-major axis is between those of the initial and target orbits. Thus, the revolution number of the Lambert solution is fewest but infeasible for a continuous low thrust. Compared with the shaping method in Ref.~\cite{zeng2017shape}, the proposed method shows its effectiveness in smaller propellant consumption ($2006.6 \, \unit{kg}$ v.s. $2596.4 \, \unit{kg}$) and shorter CPU time ($12.2\,\unit{s}$ v.s. $182.8\,\unit{s}$).

%
\begin{table}[!htb]
	\caption{Results of epochs by the PSO search}
	\label{table5}
	\vspace{0.25cm}
	\centering
	\begin{tabular}{cccccc}
		\hline \hline
		Estimation Method & $t_0,\, \unit{MJD}$ & $t_f,\,\unit{MJD} $ & $N_{\rm opt}$ & $\Delta m, \, \unit{kg}$ & CPU time, s \\ \hline
		Lambert Solution & 56483.082 & 59299.275 & 2 & 1179.047 & 0.1 \\
		Shaping Method \cite{zeng2017shape} & 56363.610 & 59000.078 & 4 & 2596.365 & 182.8 \\
		Proposed Method & 56329.586 & 59872.983 & 5 & 2006.622 & 12.2 \\
		\hline \hline
	\end{tabular}
\end{table}
%

The launch and arrival epochs are determined by the PSO search results, and the indirect method \cite{Jiang2012} is then employed to solve the corresponding propellant-optimal solutions. The techniques of the energy-propellant homotopy, costates normalization, and random guess strategy presented in Ref.~\cite{Jiang2012} are applied here and their details are omitted here for brevity. The propellant-optimal results are listed in Table~\ref{table6}, where the epochs keep the same values as those in Table~\ref{table5}. It shows that both the shaping method \cite{zeng2017shape} and proposed method provide good estimations to the revolution numbers. The optimal solution is $\Delta m_{\rm opt} = 1279.93 \, \unit{kg}$, in which the epochs are searched by the proposed method. Thus, the proposed method is superior in identifying the optimal epochs for the preliminary design.

%
\begin{table}[!htb]
	\caption{Propellant-optimal results for different epochs}
	\label{table6}
	\vspace{0.25cm}
	\centering
	\begin{tabular}{ccccc}
		\hline \hline
		Solution & $t_0,\, \unit{MJD}$ & $t_f,\,\unit{MJD} $ & $N_{\rm opt}$ & $\Delta m_{\rm opt}, \, \unit{kg}$ \\ \hline
		1 & 56483.082 & 59299.275 & 5 & 1390.679 \\
		2 & 56363.610 & 59000.078 & 4 & 1370.658 \\
		3 & 56329.586 & 59872.983 & 5 & 1279.930 \\
		\hline \hline
	\end{tabular}
\end{table}

\begin{figure}[ht!]
	\centering
	\subfigure{\includegraphics[height = 0.24\textheight]{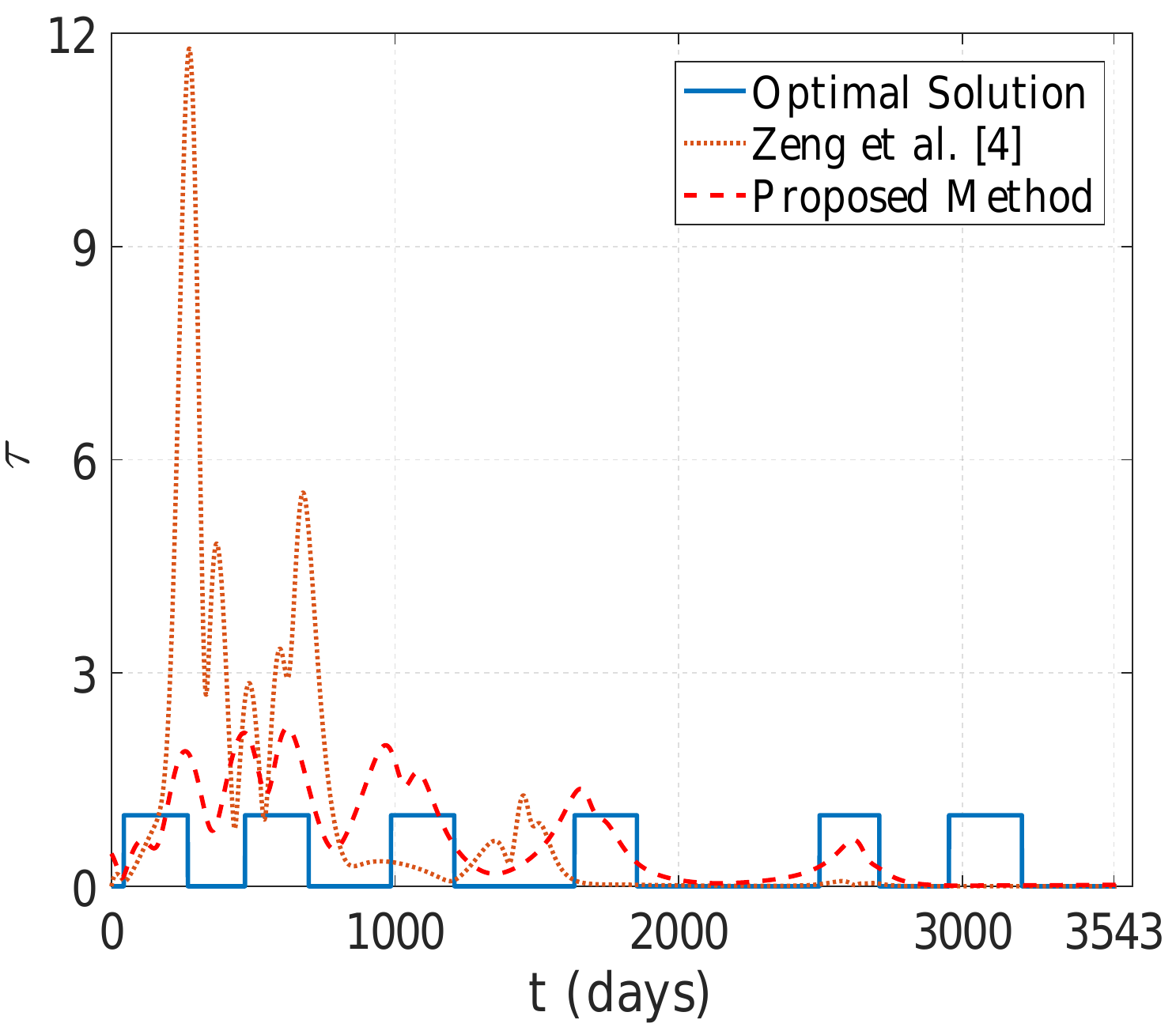}} \qquad \quad
	\subfigure{\includegraphics[height = 0.24\textheight]{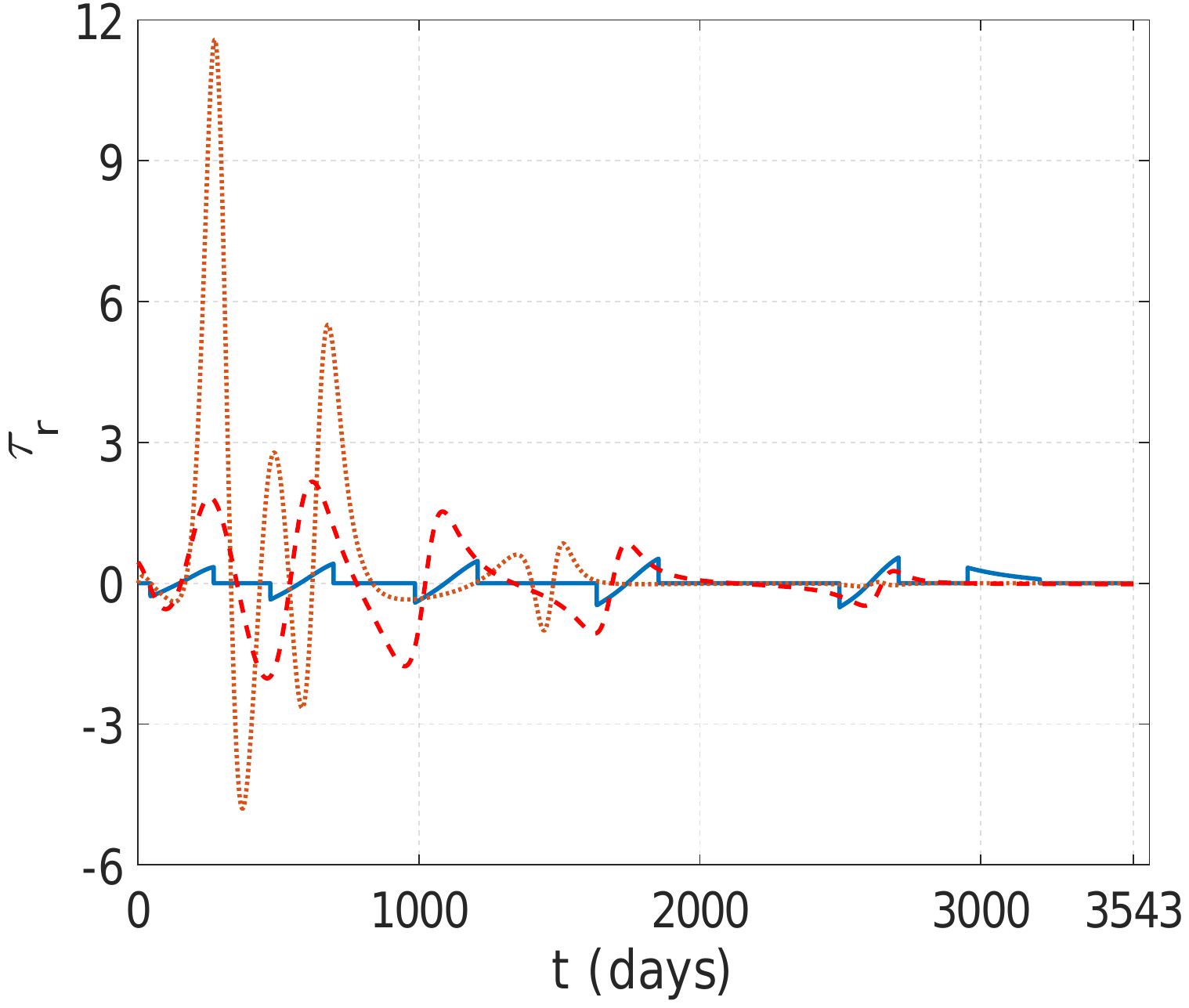}}  \\[0.4cm] 
	\subfigure{\includegraphics[height = 0.24\textheight]{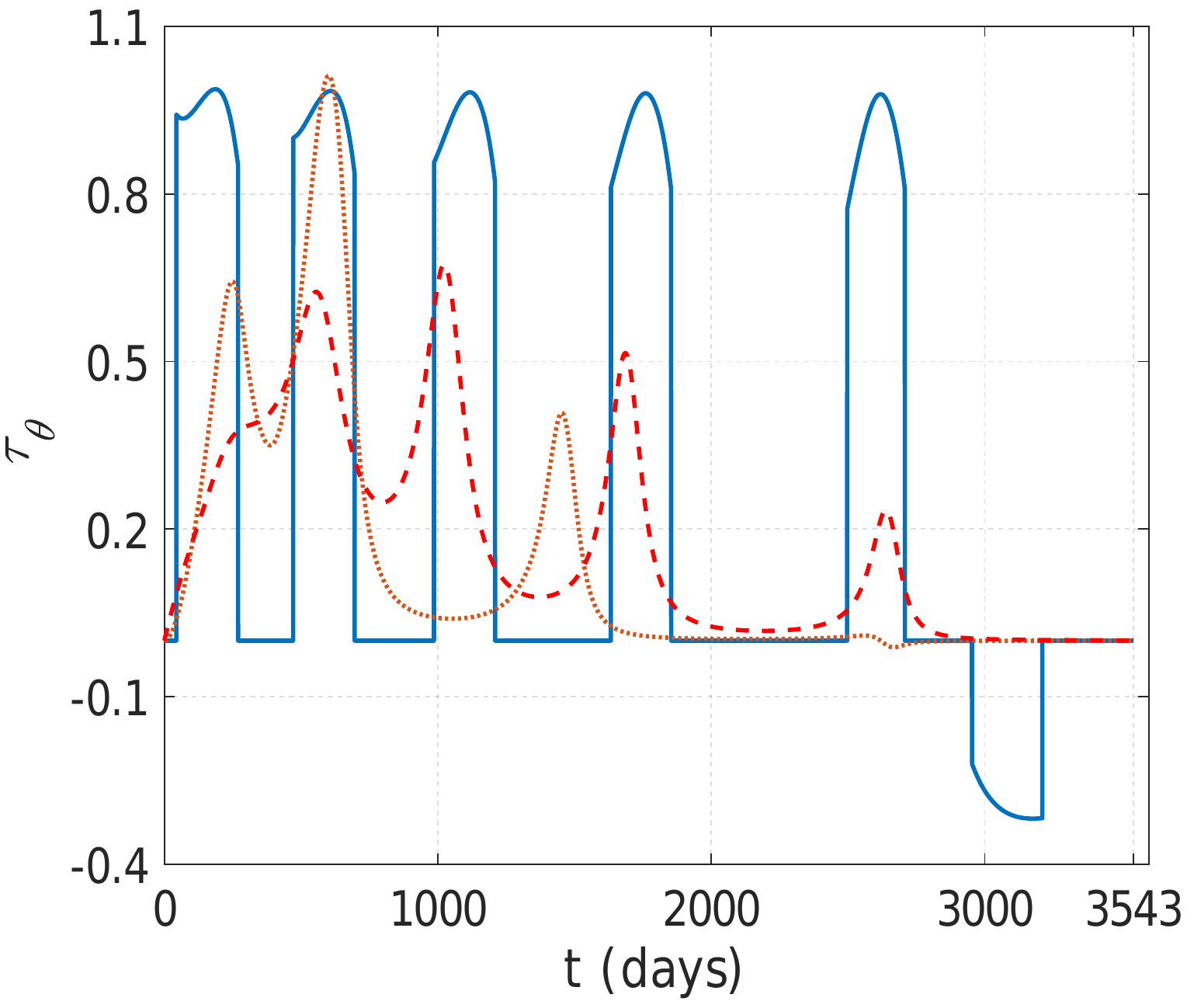}} \qquad \quad
	\subfigure{\includegraphics[height = 0.24\textheight]{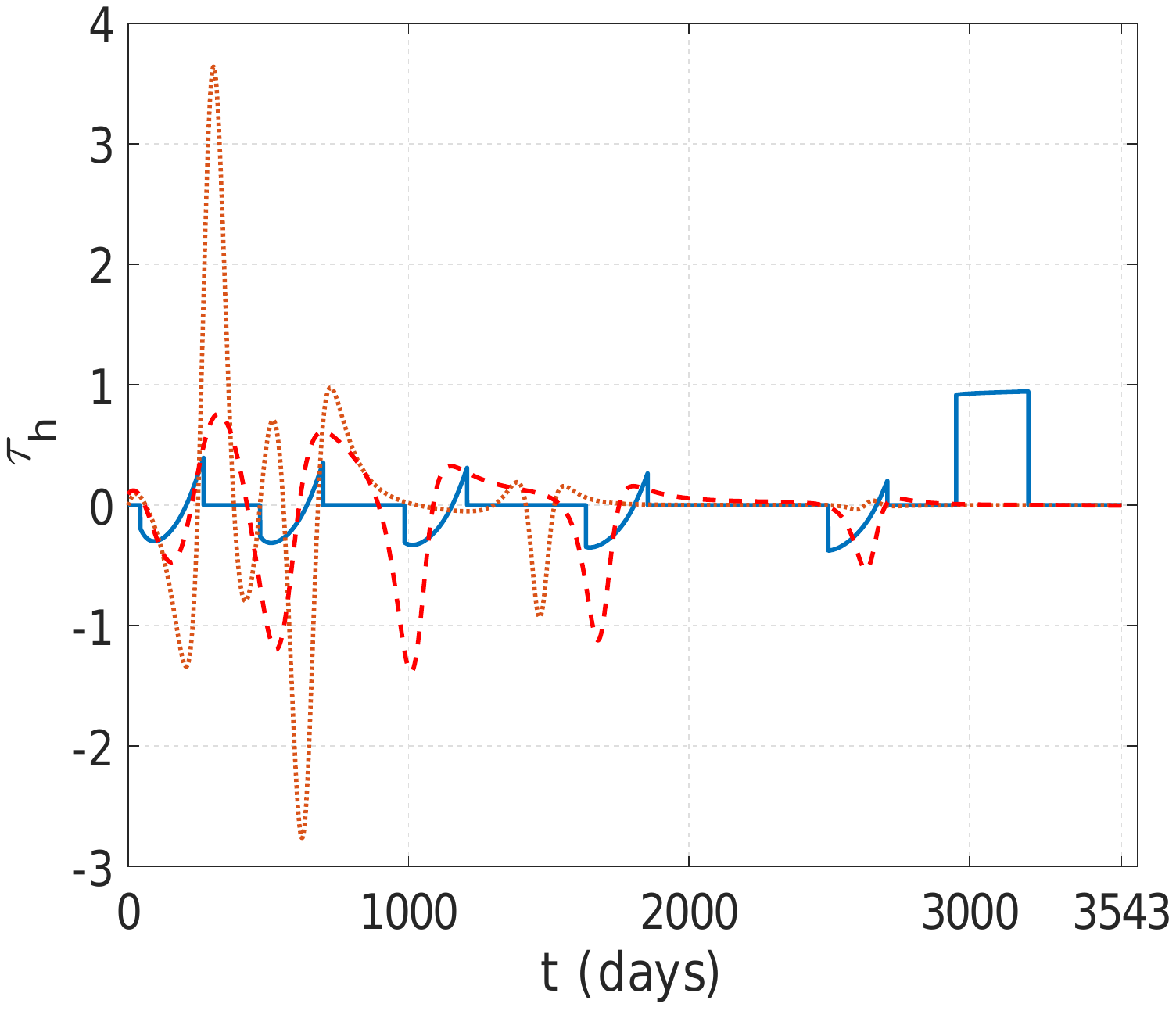}}
	\caption{Profiles of thrust components for different methods ($\V{\tau} = \V{u} \,/\, T_{\max}$).}
	\label{fig1}
\end{figure}

\begin{figure}[ht!]
	\centering
	\includegraphics[scale = 0.56]{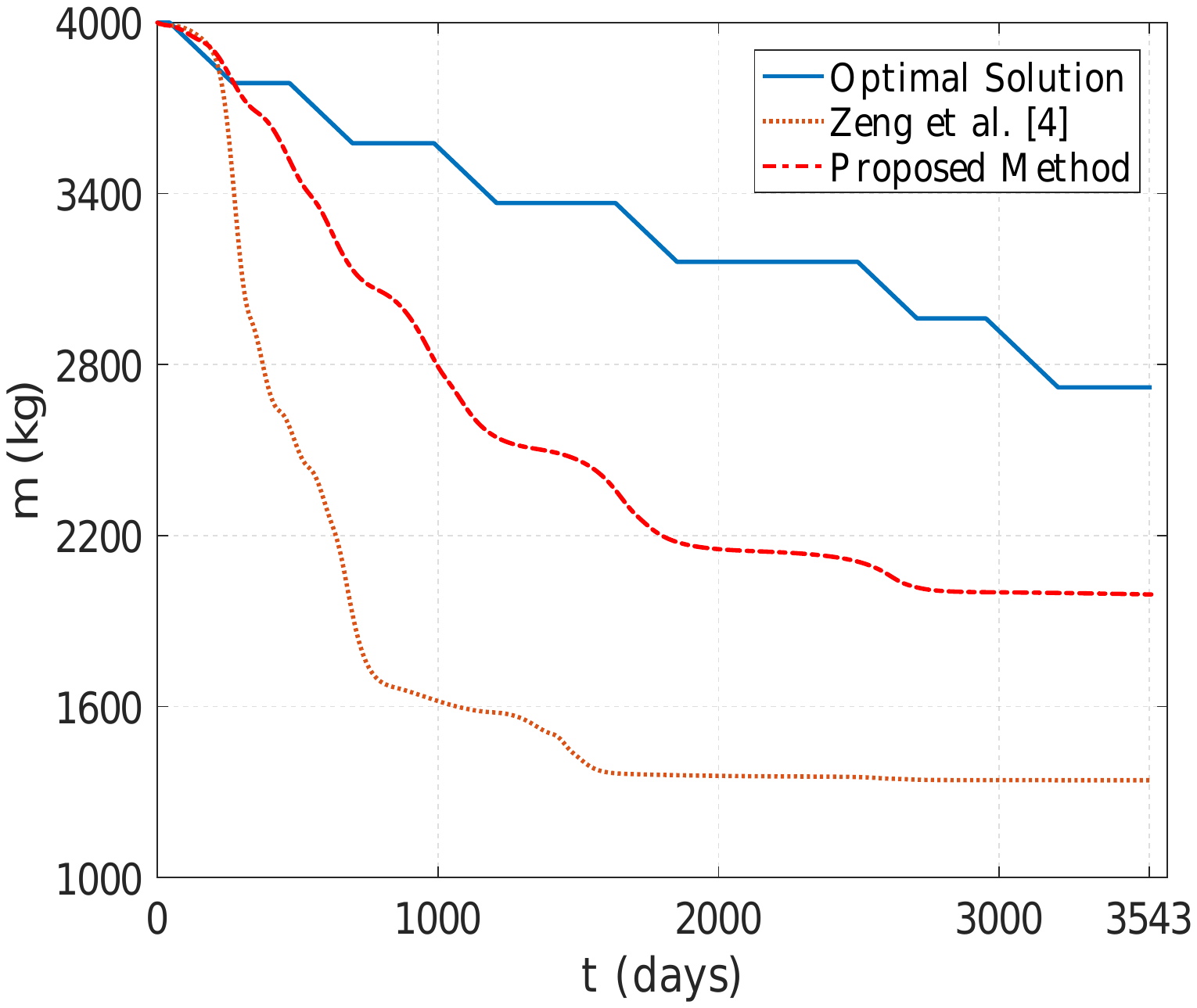}
	\caption{Histories of the spacecraft mass}
	\label{fig2}
\end{figure}

\begin{figure}[ht!]
	\centering
	\includegraphics[scale = 0.58]{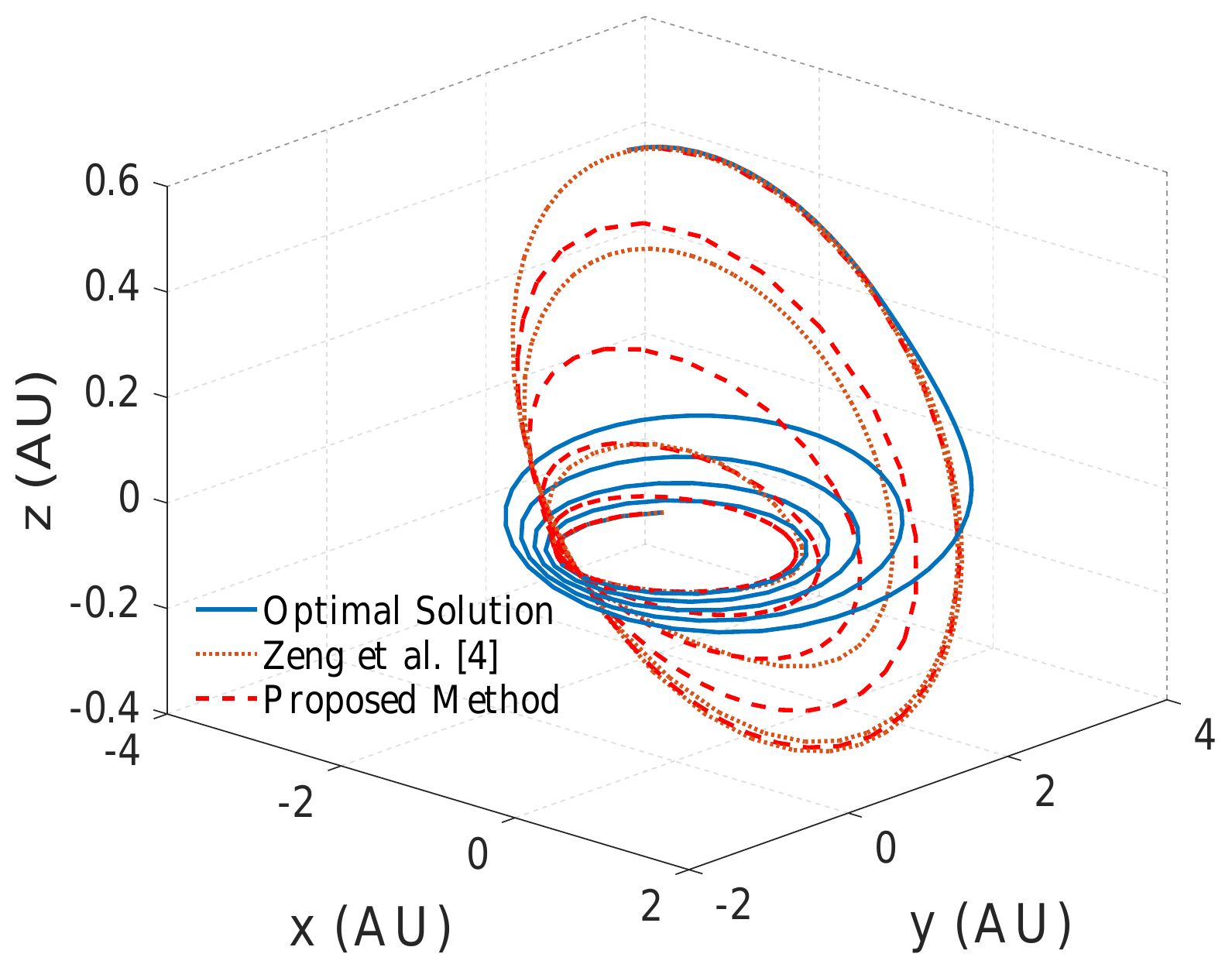}
	\caption{3-D trajectories for different methods.}
	\label{fig3}
\end{figure}

In addition, the solution 3 is investigated to demonstrate the effectiveness of the proposed method in providing initial guesses. The costates estimation technique \cite{jiang2017improving} around a shaping trajectory is used here for the energy-optimal problem, and a energy-propellant homotopy process is subsequently employed to obtain the optimal solution. For comparison, three strategies are tested: 1) random guess with the costate normalization \cite{Jiang2012}; 2) costates estimation in combination with the shaping method in Ref.~\cite{zeng2017shape}; 3) costates estimation using the proposed method. The initial costates and the percentage of converged (POC) cases out of 1000 run are summarized in Table~\ref{table7}. The proposed method improves the estimation POC to $51.4\%$, higher than the other results, \rev{and the average computational time (ACT) for one solution reduces from $12.2 \, \unit{s}$ to $1.4 \, \unit{s}$.}Furthermore, as reported in Ref.~\cite{jiang2017improving}, the performance of costates estimation relies on the assumption that the shaping trajectory is generally close to the optimal one. The profiles of the thrust components, propellant consumptions, and the trajectories of different methods are collected in Figs.~\ref{fig1}--\ref{fig3}, respectively. They show that the proposed method provides better estimations to the control and state variables.

\begin{table}[!htb]
	\caption{Results of energy-optimal problem by costates estimation}
	\label{table7}
	\vspace{0.25cm}
	\centering
	\begin{tabular}{cccc}
		\hline \hline
		Initial Guesses & Values of initial costates using MEOEs & POC, \% & \rev{ACT, s}\\ \hline
		Random Guess \cite{jiang2017improving} & - & 6.4 & \rev{12.2} \\
		Shaping method \cite{zeng2017shape} & $\left[-0.8359, \, -0.3604, \, 1.3840, \, -0.3632, -0.4956, \, 0.0137, \, 0.5\right]$ & 11.2 & \rev{9.4}\\
		Proposed method & $\left[-0.5598, \, -0.2260, \, 0.8556, \, -0.3998, -0.6346, \, 0.0067, \, 0.5\right]$  & 51.4 & \rev{1.4} \\
		Energy Optimal & $\left[-0.3580, \, -0.0319, \, 0.1597, \, -0.1689, -0.3359, \, 0.0006, \, 0.4\right]$ & - & -  \\
		\hline \hline
	\end{tabular}
\end{table}
%

%
%

%
%

%
%

\subsection{Near-Earth Asteroids Sample Return}

A sample return mission with multiple targets is studied in this subsection to design the multi-rendezvous low-thrust trajectory. The orbital elements of the Earth and targets are also given in Table~\ref{table4}. The spacecraft is assumed to have a constant specific impulse $I_{\sp}= 3000 \,\unit{s}$, a maximum thrust magnitude $T_{\max} = 0.36 \, \unit{N}$, and an initial mass $m_0 = 1500 \, \unit{kg}$. The sequential rendezvouses with 1999 $\textnormal{AO}_{10}$, 2000 $\textnormal{LG}_{6}$, and the Earth are considered in this mission \cite{chen2018multi}. After rendezvous with a target asteroid, the spacecraft takes $60 \, \unit{days}$ to carry out sampling tasks, while the sampling mass is negligible compared with the spacecraft mass. Based on the preliminary design results introduced in Ref.~\cite{chen2018multi}, the epochs when the spacecraft departs from the Earth and arrives at each targets are re-optimized here by the PSO search in combination with the proposed estimation method. 

As shown in Table~\ref{table8}, the results obtained by using the shaping method \cite{zeng2017shape} and the proposed method are searched in the range of $\Delta t \in \left[-300, \, 300 \right] \, \unit{days}$, subjected to the corresponding preliminary design results of the Lambert solution \cite{chen2018multi}. For each method, two optimal results are obtained by the indirect method, including the optimal single-leg rendezvous solution where all the epochs are fixed at the results of PSO search and the multi-leg rendezvous solution where the interior epochs are free. Thus, the propellant consumption of the second one is less. Note that the results corresponding to the Lambert solution are different from those reported in Ref.~\cite{chen2018multi} because they are different local optimal solutions. The new results obtained here have less propellant consumption. The propellant consumptions of solutions 5 and 6 are close to each other and less than others. Table~\ref{table8} clearly shows the superiority of the proposed method in searching for the globally optimal solution.

%
\begin{table}[!htb]
	\caption{Results for the multi-rendezvous mission}
	\label{table8}
	\vspace{0.25cm}
	\centering
	\begin{tabular}{cccc}
		\hline \hline
		Estimation Method & Solution & Values of epochs & $\Delta m_{\rm opt}, \unit{kg}$ \\ \hline
		\multirow{2}{*}{Lambert Solution \cite{chen2018multi}} & 1 & $\left[63608.374, \, 63960.859, \, 64752.511, \, 65336.956 \right]$ & 497.41 \\
		& 2 & $\left[63608.374, \, 63971.934, \, 64764.218, \, 65336.956 \right]$ & 490.42 \\
		\multirow{2}{*}{Shaping method \cite{zeng2017shape}} & 3 & $\left[63451.125, \, 64135.812, \, 64725.645, \, 65420.367\right]$ & 616.34 \\
		& 4 & $\left[63451.125, \, 63948.783, \, 64767.835, \, 65420.367\right]$ & 451.34 \\
		\multirow{2}{*}{Proposed method} & 5 & $\left[63309.826, \, 63897.212, \, 64762.270, \, 65337.164 \right]$ & 442.64 \\
		& 6 & $\left[63309.826, \, 63897.375, \, 64766.455, \, 65337.164 \right]$ & 442.20 \\
		\hline \hline
	\end{tabular}
\end{table}
%

\begin{figure}[ht!]
	\centering
	\includegraphics[scale = 0.56]{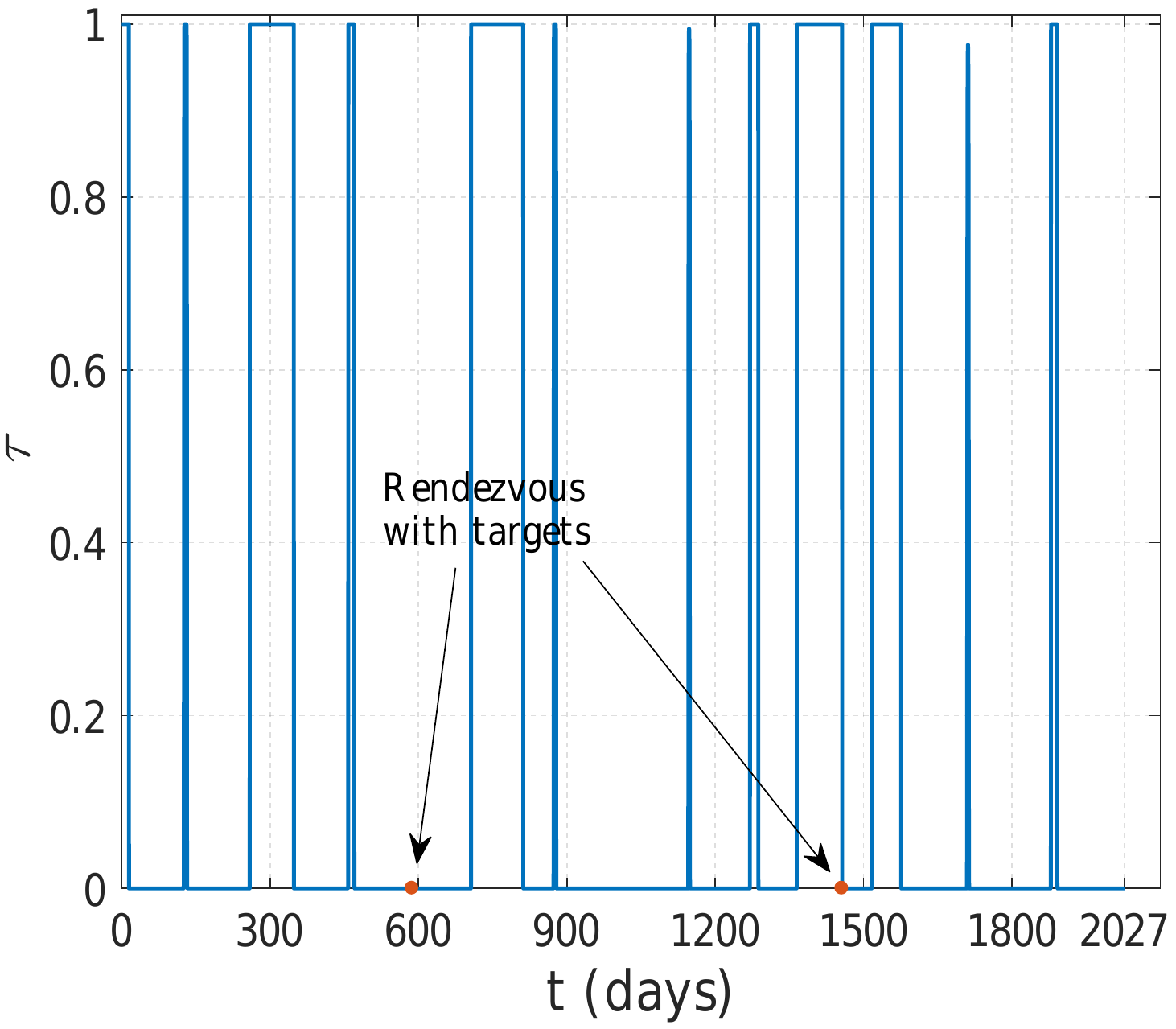}
	\caption{Thrust magnitude for the multi-rendezvous mission (solution 6).}
	\label{fig4}
\end{figure}

\begin{figure}[ht!]
	\centering
	\includegraphics[scale = 0.56]{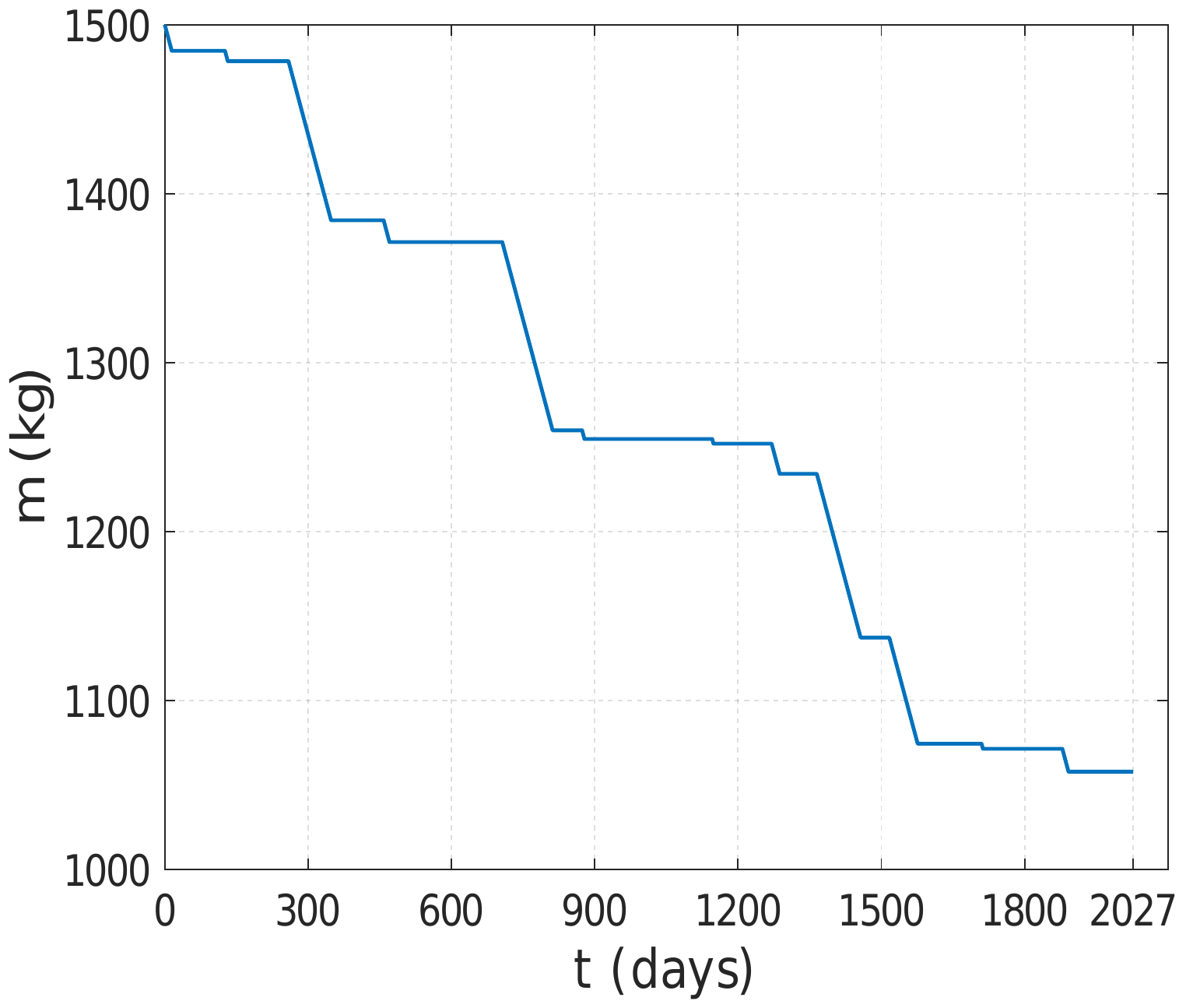}
	\caption{Histories of the spacecraft mass for the multi-rendezvous mission.}
	\label{fig5}
\end{figure}

\begin{figure}[ht!]
	\centering
	\includegraphics[scale = 0.56]{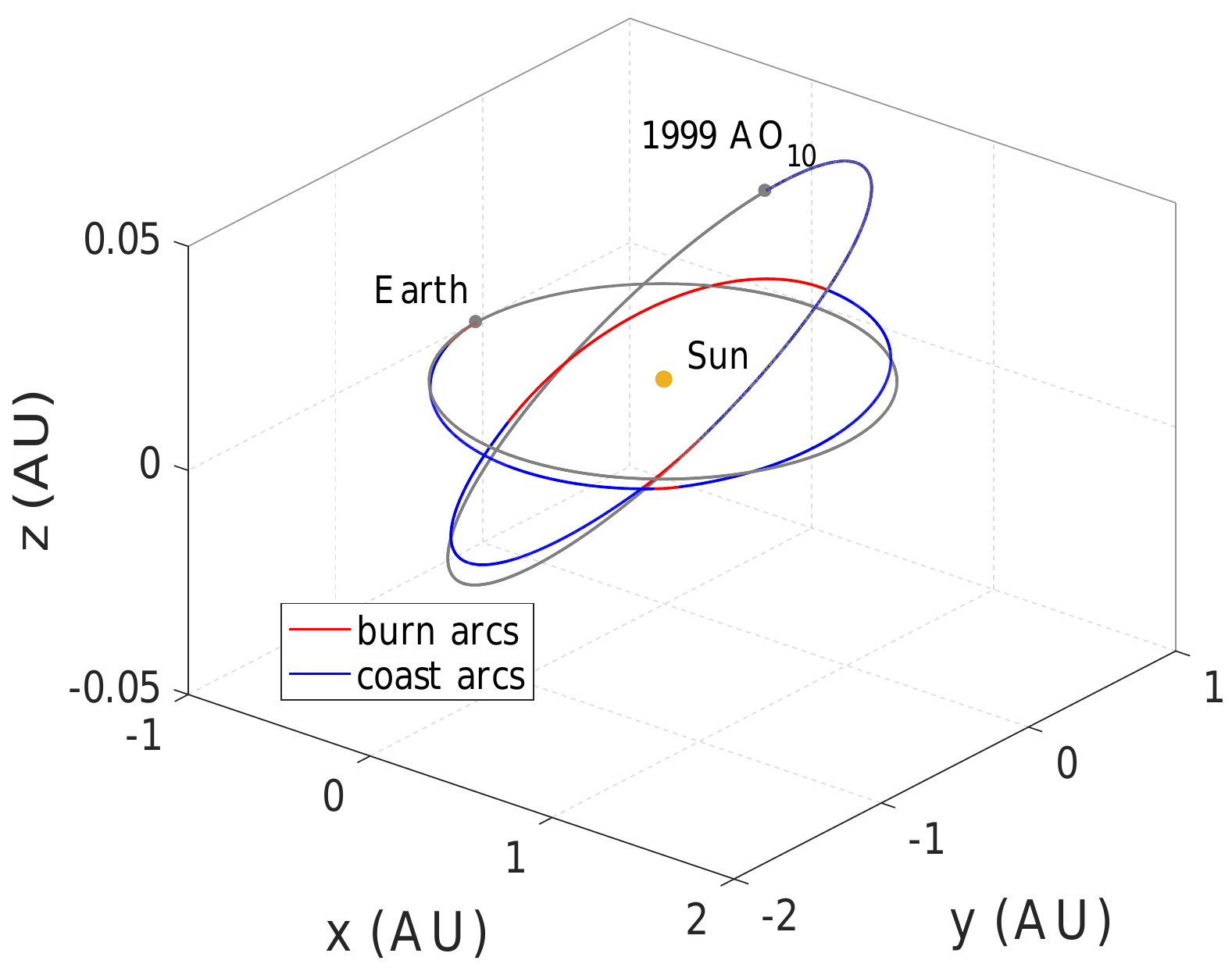}
	\caption{Propellant-optimal trajectory from Earth to 1999 $\textnormal{AO}_{10}$.}
	\label{fig6}
\end{figure}

\begin{figure}[ht!]
	\centering
	\includegraphics[scale = 0.56]{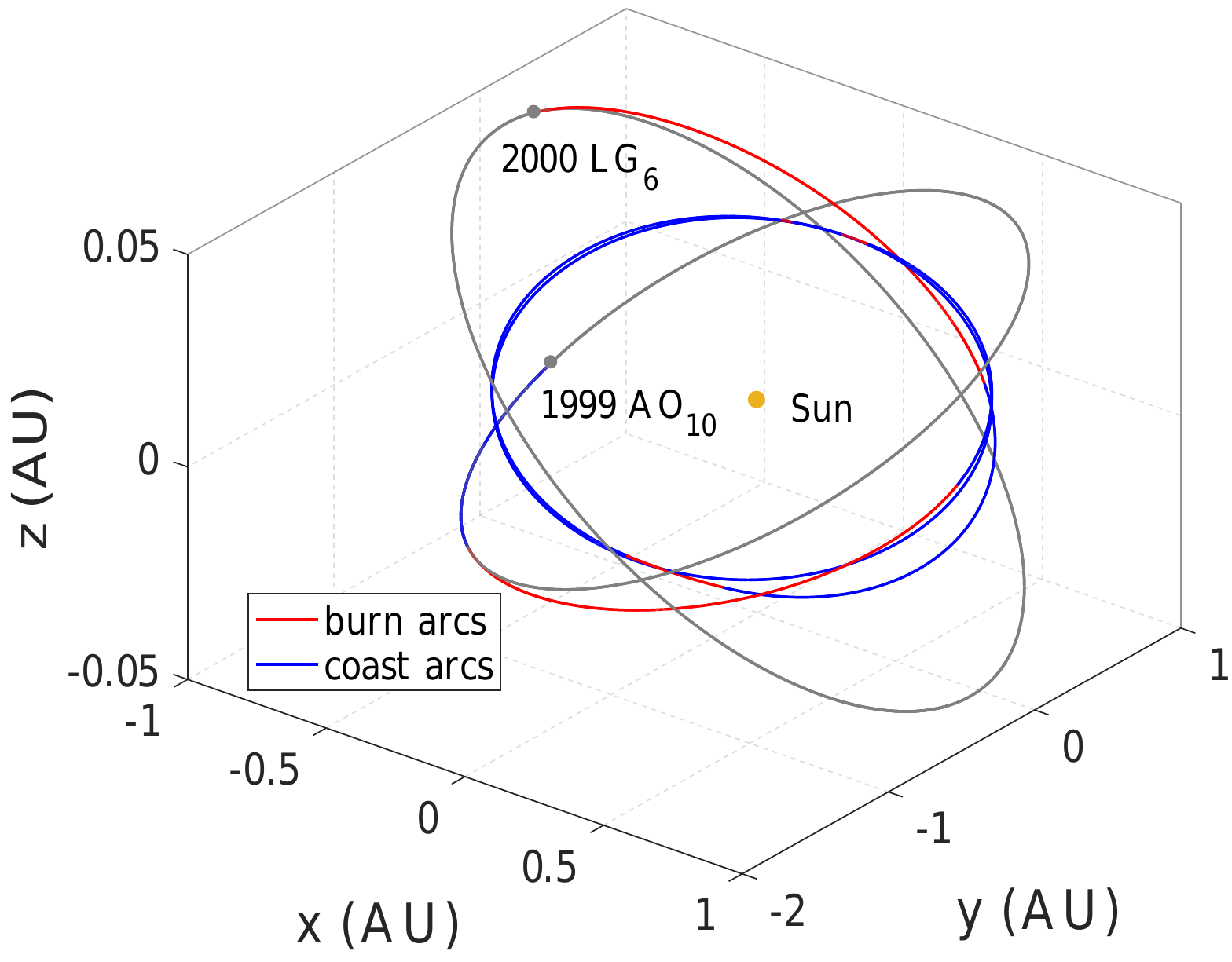}
	\caption{Propellant-optimal trajectory from 1999 $\textnormal{AO}_{10}$ to 2000 $\textnormal{LG}_6$.}
	\label{fig7}
\end{figure}

\begin{figure}[ht!]
	\centering
	\includegraphics[scale = 0.56]{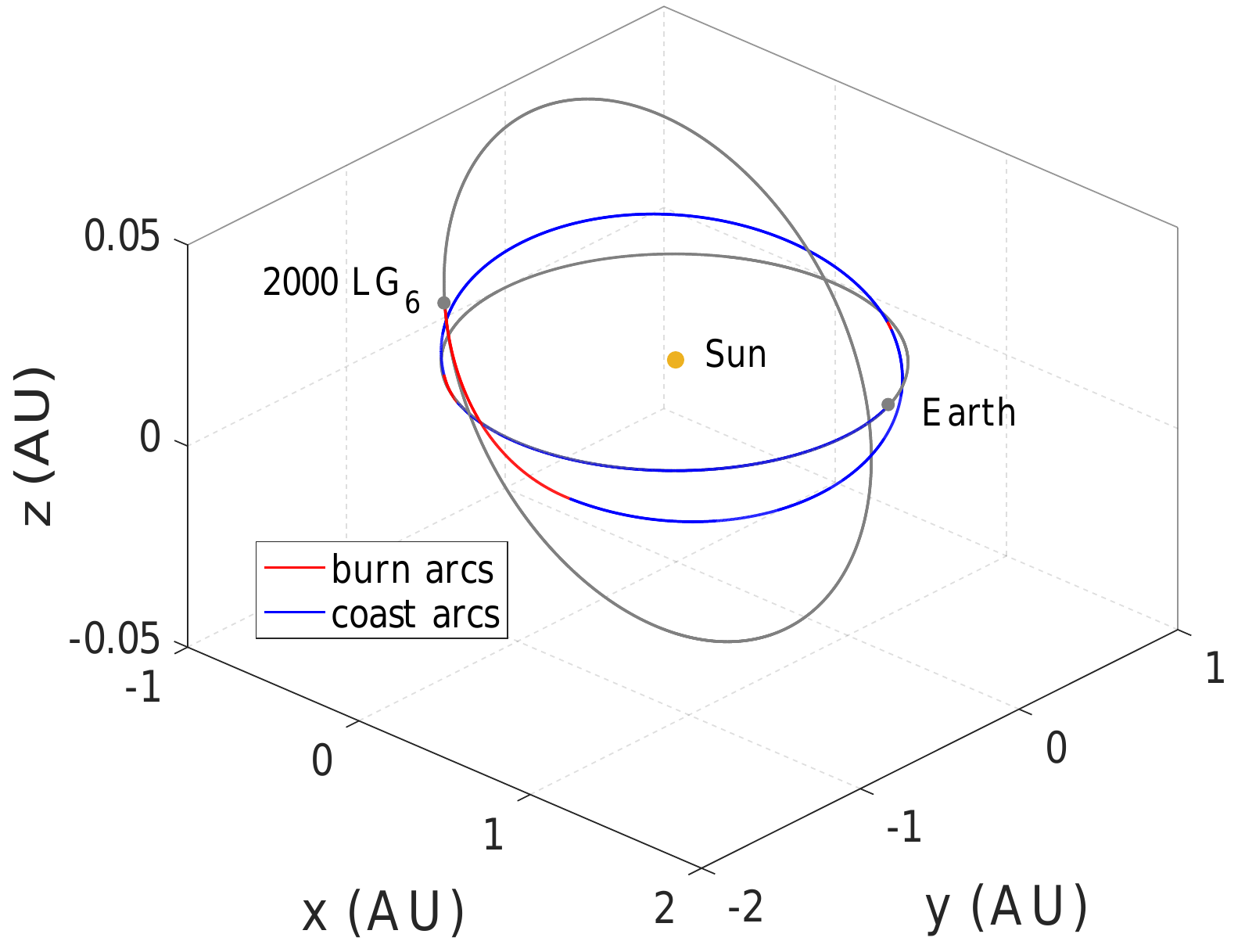}
	\caption{Propellant-optimal trajectory from 2000 $\textnormal{LG}_6$ to Earth.}
	\label{fig8}
\end{figure}

Furthermore, the optimal solution of the multi-rendezvous mission takes $\Delta m_{\rm opt} = 442.20 \, \unit{kg}$ in total. The spacecraft firstly departs from the Earth at $63309.826 \, \unit{MJD}$ to rendezvous with the 1999 $\textnormal{AO}_{10}$ in $\Delta t_1 = 587.549 \, \unit{days}$ with a propellant consumption $\Delta m_1 = 128.64 \, \unit{kg}$. Then, the second leg from the 1999 $\textnormal{AO}_{10}$ to 2000 $\textnormal{LG}_{6}$ takes $\Delta t_2 = 809.080 \, \unit{days}$ and $\Delta m_2 = 234.19 \, \unit{kg}$. Finally, the spacecraft returns to the Earth with another $\Delta t_3 = 570.709 \, \unit{days}$ and $\Delta m_3 = 79.37 \, \unit{kg}$. The histories of the thrust magnitude and spacecraft mass are plotted in Figs.~\ref{fig4} and~\ref{fig5}, respectively. The structure of bang-bang control is obtained, in which several burn arcs are very short (the lengths of the three shortest arcs are all less than $8\, \unit{days}$). The propellant-optimal rendezvous trajectories are plotted in Figs.~\ref{fig6}--\ref{fig8}. The spacecraft arrives at the 1999 $\textnormal{AO}_{10}$ and Earth earlier than the given rendezvous epochs, implying that this solution can be further improved by adjusting the launch and return epochs.

%
%

%
%

%
%

%
%

%
%

\section{Conclusions}

In this paper, an analytical shaping method is proposed to approximate the general 3-D multi-revolution low-thrust rendezvous trajectory using the cubic spline functions. The boundary constraints are all analytically satisfied with the parameters of the shape function. When neglecting the thrust magnitude constraint, the Newtonian iterative process is no longer needed, and a fast and robust shaping algorithm is achieved. Moreover, a parameter optimization problem with several discrete segments and constrained points is formulated and solved to provide better approximation results considering the constraints on the physical parameters of the spacecraft and its engine.

Compared with the existing 3-D shaping method, the proposed method shows remarkable advantages both in the less total velocity increment and smaller maximum acceleration. The effectiveness and robustness of the proposed method are demonstrated via a multi-revolution interplanetary rendezvous mission and a multi-rendezvous sample return mission. The proposed method provides good estimations for the global search and improves the convergence percentage of the initial guesses for the indirect method.

\section{Acknowledgment}
This work was supported by the National Natural Science Foundation of China (grant nos. 12022214).

\section*{Appendix: Expressions for $\diff \V{r} / \diff {\xoe}$, $\diff \V{r}^2 / \diff^2 {\xoe}$, and $\diff {r} / \diff {\xoe}$}

In this section, the derivatives in terms of the MEOEs for the shaping approximation are presented. For clarity, the following parameters are defined:
\begin{align*}
&\eta_1 = 1 + f\cos L + g\sin L, \quad \eta_2 = \left(f\sin L - g\cos L\right)/\eta_1, \quad \eta_3 = \left(f\cos L + g\sin L\right)/\eta_1, \\[0.3cm]
&\eta_4 = h\sin L -k\cos L, \quad \eta_5 = h\cos L + k\sin L, \quad \eta_6 = h\cos L - k\sin L, \\[0.3cm]
&\eta_7 = \sin L - \eta_2 \cos L, \quad \eta_8 = \cos L + \eta_2 \sin L, \quad \eta_9 = 2\,\eta_2^2 + \eta_3 - 1
\end{align*}
According to Eq.~\eqref{sr}, the derivatives $\diff \V{r} / \diff \xoe $ are obtained
\begin{align*}
&\frac{\partial \V{r}}{\partial p} = \frac{\V{r}}{p}, \quad \frac{\partial \V{r}}{\partial f} = -\frac{\V{r}\cos L}{\eta_1}, \quad \frac{\partial \V{r}}{\partial g} = -\frac{\V{r}\sin L}{\eta_1},  \\[0.3cm]
&\frac{\partial \V{r}}{\partial h} = -\frac{2 \, \V{r} \, h}{\beta^2} + \frac{2 \, r}{\beta^2} \left[\eta_5, \, -\eta_4, \, \sin L\right]^{\T}, \quad \frac{\partial \V{r}}{\partial k} = -\frac{2 \, \V{r} \, k}{\beta^2} + \frac{2 \, r}{\beta^2} \left[\eta_4, \, \eta_5, \, -\cos L\right]^{\T}, \\[0.3cm]
&\frac{\partial \V{r}}{\partial L} = \eta_2 \, \V{r} + \frac{r}{\beta^2} \left[2\,h\,k\cos L - \left(1+\alpha^2\right)\sin L, \, \left(1-\alpha^2\right)\cos L - 2\,h\,k\sin L, \, 2\,\eta_5\right]^{\T}
\end{align*}
Then, the derivatives $\diff \V{r}^2 / \diff^2 {\xoe}$ are derived as
\begin{align*}
&\frac{\partial^2 \V{r}}{\partial p^2} = 0, \;\; \frac{\partial^2 \V{r}}{\partial p \, \partial f} = \frac{\partial \V{r}}{p \, \partial f}, \;\; \frac{\partial^2 \V{r}}{\partial p \, \partial g} = \frac{\partial \V{r}}{p \, \partial g}, \;\;  \frac{\partial^2 \V{r}}{\partial p \, \partial h} = \frac{\partial \V{r}}{p \, \partial h}, \;\; \frac{\partial^2 \V{r}}{\partial p \, \partial k} = \frac{\partial \V{r}}{p \, \partial k}, \;\; \frac{\partial^2 \V{r}}{\partial p \, \partial L} = \frac{\partial \V{r}}{p \, \partial L}, \\[0.3cm]
&\frac{\partial^2 \V{r}}{\partial f^2} = \frac{2\,\V{r} \cos^2 L}{\eta_1^2}, \quad \frac{\partial^2 \V{r}}{\partial h^2} = \frac{8\,h^2 \, \V{r}}{\beta^4} - \frac{4\,r}{\beta^4}\left[2\,h\,\eta_5 + k\,\eta_4, \, \sin L + k\,\eta_5 - 2\,h\,\eta_4, \, 2\, h \sin L + \eta_4 \right]^{\T}, \\[0.3cm]
&\frac{\partial^2 \V{r}}{\partial g^2} = \frac{2\,\V{r} \sin^2 L}{\eta_1^2}, \quad \frac{\partial^2 \V{r}}{\partial k^2} = \frac{8\,k^2 \,\V{r}}{\beta^4} - \frac{4\,r}{\beta^4}\left[\cos L + h\,\eta_5 + 2\,k\,\eta_4, \, 2\,k\,\eta_5 - h\,\eta_4, \, \eta_4-2\, k \cos L\right]^{\T}, \\[0.3cm]
&\frac{\partial^2 \V{r}}{\partial L^2} = \eta_9 \, \V{r} + \frac{2\,r\,\eta_2}{\beta^2}\left[ 2 \, h \, k \cos L - \left(1+\alpha^2\right) \sin L, \, \left(1-\alpha^2\right) \cos L - 2 \, h \, k\sin L, \, {2 \, \eta_5}\right]^{\T}, \\[0.3cm]
&\frac{\partial^2 \V{r}}{\partial f \, \partial h} = -\frac{\cos L}{\eta_1} \frac{\partial \V{r}}{\partial h}, \quad \frac{\partial^2 \V{r}}{\partial f \, \partial k} = -\frac{\cos L}{\eta_1} \frac{\partial \V{r}}{\partial k}, \quad \frac{\partial^2 \V{r}}{\partial g \, \partial h} = -\frac{\sin L}{\eta_1} \frac{\partial \V{r}}{\partial h}, \quad \frac{\partial^2 \V{r}}{\partial g \, \partial k} = -\frac{\sin L}{\eta_1} \frac{\partial \V{r}}{\partial k}, \\[0.3cm]
&\frac{\partial^2 \V{r}}{\partial f \, \partial g} = \frac{\V{r} \sin 2L}{\eta_1^2}, \quad \frac{\partial^2 \V{r}}{\partial h \, \partial k} = \frac{8 \, h \, k \, \V{r}}{\beta^4} + \frac{2\,r}{\beta^4}\left[\left(2 - \beta^2\right) \sin L, \, \left(2 - \beta^2\right) \cos L, \, 2\,\eta_6 \right]^{\T}, \\[0.3cm]
&\frac{\partial^2 \V{r}}{\partial f\, \partial L} = \frac{\V{r} \, \eta_7 }{\eta_1}-\frac{\cos L}{\eta_1}\frac{\partial \V{r}}{\partial L}, \quad \frac{\partial^2 \V{r}}{\partial h \, \partial L} = - \frac{2 \, h}{\beta^2}\frac{\partial \V{r}}{\partial L} + \frac{2 \, r}{\beta^2}\left[\eta_2 \, \eta_5 -\eta_4, \, -\eta_2 \, \eta_4 -\eta_5, \, \eta_8 \right]^{\T},  \\[0.3cm]
&\frac{\partial^2 \V{r}}{\partial g\, \partial L} = -\frac{\V{r} \, \eta_8 }{\eta_1}-\frac{\sin L}{\eta_1}\frac{\partial \V{r}}{\partial L}, \quad \frac{\partial^2 \V{r}}{\partial k \, \partial L} = - \frac{2 \,k}{\beta^2}\frac{\partial \V{r}}{\partial L} + \frac{2 \, r}{\beta^2} \left[\eta_2 \, \eta_4 + \eta_5, \, \eta_2 \, \eta_5 - \eta_4, \, \eta_7\right]^{\T} 
\end{align*}
Since $r = \left\| \V{r}\right\| = p\,/\left(1+f\cos L + g\sin L\right)$, the derivative $\diff {r} / \diff {\xoe}$ is derived as
\begin{equation*}
\frac{\partial r}{\partial \xoe} = \left[\frac{r}{p}, \, -\frac{r\cos L}{\eta_1}, \, -\frac{r\sin L}{\eta_1}, \, 0, \, 0, \, r \, \eta_2\right]^{\T}
\end{equation*}
%


\bibliographystyle{elsart-num_doi}                          
\bibliography{biblio_di}

\end{document}